\documentclass[letterpaper,twocolumn,10pt]{article}
\usepackage{usenix}

\usepackage{float}
\usepackage{cite}
\usepackage[normalem]{ulem}
\useunder{\uline}{\ul}{}
\usepackage{amsmath,amssymb,amsfonts}
\usepackage{algorithmic}
\usepackage{graphicx}
\usepackage{textcomp}
\usepackage{bmpsize}
\usepackage{xcolor}
\usepackage{lipsum}
\usepackage{amsthm}
\usepackage{caption}
\usepackage{bbm}
\usepackage{wasysym}
\usepackage{booktabs}
\usepackage{multirow}
\usepackage{enumitem}
\usepackage{subcaption}
\usepackage{threeparttable} 
\usepackage{tikz}
\usepackage{fontawesome}
\usepackage{hyperref}
\def\BibTeX{{\rm B\kern-.05em{\sc i\kern-.025em b}\kern-.08em
    T\kern-.1667em\lower.7ex\hbox{E}\kern-.125emX}}

\newcommand{\notcheckmark}{%
  \ooalign{%
    $\checkmark$\cr
    \hidewidth\hspace{0.2em}\raisebox{1ex}{\rule{0.6em}{0.09ex}}\hidewidth\cr
  }%
}

\newtheorem{definition}{Definition}

\usepackage{filecontents}

\begin{document}

\date{}

\title{\Large \bf SoK: Rethinking Membership Inference in Tabular Machine Learning}


\author{Cristina Pêra$^{3,4}$, Tânia Carvalho$^{2}$, Maxime Cordy$^{2}$, Luís Antunes$^{1,3}$ \\ $^{1}$Department of Computer Science, Faculty of Sciences of the University of Porto, Portugal \\ $^{2}$SnT, University of Luxembourg, Luxembourg \\ $^{3}$TekPrivacy, Portugal \\ $^{4}$Corresponding author: cpera@tekprivacy.com }

\maketitle

\begin{abstract}
Membership Inference Attacks (MIAs) are currently the dominant approach to evaluating privacy in machine learning applications. Despite their significance, several concerns remain unexplored, particularly with regard to tabular data. Existing evaluations focus predominantly on high-dimensional datasets, leaving the privacy behavior of small-to-medium datasets unexplored.
In this paper, we present a new taxonomy for MIAs and conduct a comprehensive study of the factors influencing vulnerability. We evaluate four distinct state-of-the-art attacks across diverse tabular datasets and model architectures.
Our results show that datasets with higher class granularity and majority class percentage are more susceptible to MIAs. We further demonstrate that advanced MIAs are highly dependent on adversarial knowledge of the target model, while reference model-free attacks often achieve statistically comparable results. 
Most importantly, we show that records with unique signatures (single-outs) remain highly vulnerable, even when common defenses are applied. These findings indicate that current defenses reduce average leakage, but fail to adequately protect the most sensitive individuals.
\end{abstract}


\section{Introduction}
In the past decade, Artificial Intelligence (AI) has reached astonishing progress. Currently, nearly every aspect of daily life involves at least one AI-driven system, for instance, personalized recommendations, virtual assistants, and autonomous vehicles. Although such technologies automate processes and improve convenience and efficiency, their rapid development has also raised societal concerns, particularly regarding privacy, as personal information is increasingly stored and reused. 
This underscores the importance of developing tools and methodologies that can effectively support both developers and existing systems in evaluating their compliance.

A quantitative privacy measure that is currently on the spotlight is Membership Inference Attack (MIA)~\cite{Shokri-MIA-Against-ML,pyrgelis2017knock,yan2022membership}. This is, so far, the most widely used and easiest to interpret approach. Put simply, MIA determines whether a specific record (i.e. an individual) is part of the training dataset. For example, if a model was trained exclusively on patients from a mental health clinic, successfully inferring that a person's record was part of the training set reveals that the individual received psychological treatment. Such a disclosure exposes highly sensitive personal information while also compromising the confidentiality and reputation of the clinic. 
Therefore, MIAs are valuable tools for auditing data privacy and regulatory compliance of machine learning (ML) models. For instance, under the legal requirement of the \textit{right to be forgotten} (Art. 17, GDPR), MIAs can verify if the model does not include the specific data point following a deletion request.

This type of attack has been explored on a wide variety of data types and learning tasks. Initially, MIAs were studied in the context of classification problems~\cite{li2021membership,Shokri-MIA-Against-ML,Hierarchical-tree-based,MIA-first-prin,Salem-ML-Leaks,rmia,Enhanced-MIA-Game}. Since then, research has been extended to other domains such as regression~\cite{gupta2021membership}, recommendation systems~\cite{zhang2021membership,chi2024shadow,he2025membership}, graph learning~\cite{wu2021adapting}, natural language processing~\cite{shejwalkar2021membershipnlp}, and even wireless signal analysis~\cite{shi2022membership} and location data~\cite{pyrgelis2017knock}. Recently, attention has shifted towards MIAs in Large Language Models (LLMs)~\cite{meeus2025sok,wu2025membership,he2025towards,wen2024membership} and Tabular Foundation Models~\cite{carvalho2026privacy}. 

Although multiple data domains have been explored for MIAs, tabular data remain overlooked. Structured data are the predominant data format in real-world applications, such as healthcare, finance, demographics, and government statistics. 
The nature of the attributes and the presence of sensitive information in tabular data make privacy analysis essential. Yet most existing MIA evaluations neglect this setting, particularly simple tabular datasets.

\begin{figure}[ht!]
    \centering
    \includegraphics[width=0.9\columnwidth]{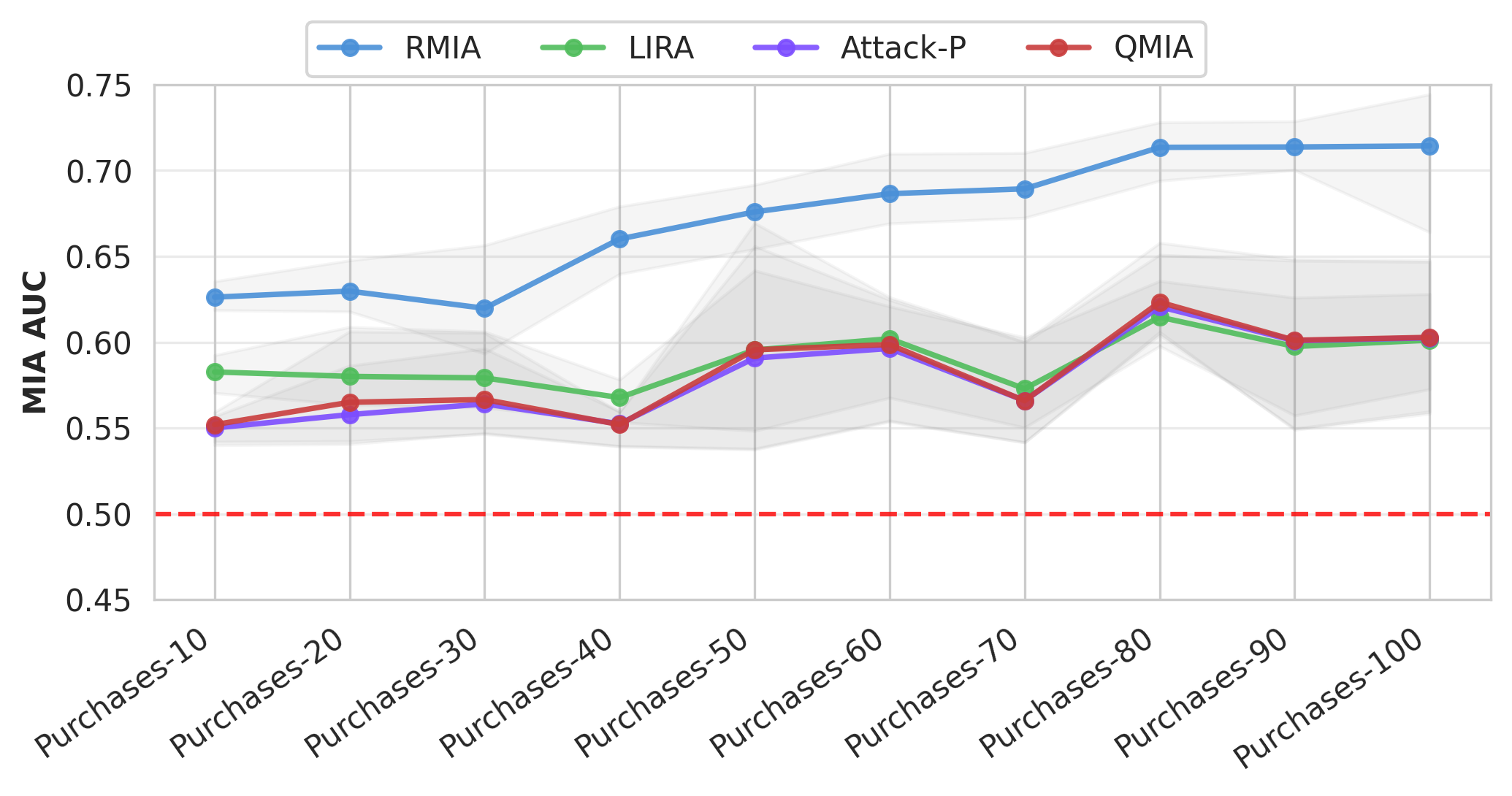}
    \caption{Membership Inference Attacks effectiveness for different Purchases variants. Results are reported for five runs.}
    \label{fig:purchases}
\end{figure}

In this work, we focus on MIAs in supervised classification, one of the most widely adopted ML paradigms in practice. We identify three key research gaps.
First, models trained on simple tabular data are rarely evaluated, making MIAs less reliable in such data settings. 
In particular, the relationship between essential dataset characteristics and MIA effectiveness is largely unexplored. Figure~\ref{fig:purchases} illustrates this challenge: although larger training datasets are generally expected to improve privacy by reducing overfitting, increasing class granularity consistently leads to higher MIA performance.
Second, while model-based attacks are often considered the most effective approach, they require surrogate models, substantial computational resources, and assumptions about the target model that may not be applicable in practice. It remains unclear whether these additional costs consistently translate into better attack performance than simpler reference model-free alternatives. Third, the privacy risks faced by single-outs, namely individuals with highly unique records~\cite{k-anonymity}, have received no attention in the MIA literature, despite being the individuals most likely to suffer severe privacy breaches.

\textbf{Overall, we make the following contributions.}

\begin{enumerate}[label=\roman*)]
    \item We conduct thorough investigations into existing research on MIAs and their respective defenses.
     Based on our finding, we suggest a new taxonomy for MIAs according to the type of adversary access instead of knowledge. The systematization leads to several key insights: \textit{i)} most existing MIAs incur high computational costs due to the need to train additional ML models; \textit{ii)} the majority are designed for deep learning, with relatively few applicable to tabular data; \textit{iii)} many attacks are not evaluated against defense mechanisms, potentially overestimating their effectiveness and limiting their practical assessment; \textit{iv)} some defenses (e.g., MemGuard~\cite{MemGuard}, Nirvana~\cite{White-Box-MIA-DL}) are evaluated against only a single MIA, leaving their robustness across different attack instances and methods uncertain; and \textit{v)} several proposed attacks do not provide publicly available code, hindering reproducibility (see Table~\ref{tab:summary}).
     
    \item We comprehensively evaluate four state-of-the-art MIAs across ten small-to-medium tabular datasets spanning diverse characteristics. To reflect the nature of these datasets, we evaluate a variety of ML models 
    , including simpler architectures which are frequently employed in this scenario, yet often overlooked in MIA studies. 
    Our findings provide practical guidance for identifying privacy risks early in the ML pipeline, enabling informed preprocessing decisions, such as balancing target entropy and majority class percentage, or reducing highly granular class structures, as well as selecting models that are less susceptible to MIAs.

    \item We demonstrate that effective privacy auditing can be performed with simpler and more resource-efficient methods without compromising attack effectiveness. In particular, model-free MIAs achieve statistically comparable performance to model-based attacks, despite requiring fewer assumptions about the target model, which better reflects realistic black-box scenarios where little or no knowledge of the target model is available.
    
    \item Finally, we investigate the vulnerability of single-outs to MIAs. 
    Due to their distinctive attribute combinations, such records are more likely to be retained by ML models, increasing the adversary's confidence. Our results show that single-outs are consistently more vulnerable than non-single-outs and remain highly susceptible even when defense mechanisms are applied. To mitigate this risk, we evaluate prior pseudonymization of the training data and demonstrate that it can reduce attack effectiveness while preserving predictive utility.

\end{enumerate}


\textbf{Tabular MIAs as a SoK.}
Several surveys have reviewed MIAs~\cite{niu2024survey,hu2023defenses,niu2025comparing,meeus2025sok}. Our SoK complements these works by offering a practical overview for researchers and practitioners. We introduce a taxonomy based on the minimum adversarial access required, rather than the traditional white/black-box distinction, avoiding overlapping assumptions and enabling clearer comparisons between attack families.

Furthermore, current MIA research mainly focuses on image datasets and complex tabular benchmarks with hundreds of classes, where deep learning models are typically employed (e.g.,~\cite{Yeom,Enhanced-MIA-Game,rmia,MIA-first-prin,tao2025range}). Consequently, MIAs on simple-to-medium tabular classification tasks remains poorly understood. We address this gap through a comprehensive empirical evaluation across diverse tabular datasets. 

\textbf{Systematization Scope and Roadmap.} 
We conducted a Google Scholar search using combinations of keywords related to MIAs, privacy-preserving ML, defense for MIAs and tabular data. We considered works published since 2017, excluding prompt-based LLM classification approaches.
Among the retrieved literature, we identified 33 representative MIA methods published in recognized peer-reviewed venues that are sufficiently specified through algorithmic descriptions and/or publicly available implementations. The remaining of the paper is organized as follows. Section~\ref{sec:foudations} introduces the preliminaries of the MIA game. Section~\ref{sec:sota} reviews and discusses existing MIA literature, followed by defenses in Section~\ref{sec:defenses_sota} and evaluation metrics in Section~\ref{sec:miaeval}. Sections~\ref{sec:mia_tabular} and~\ref{sec:singleouts} evaluate MIA effectiveness and single-out vulnerability, respectively. Section~\ref{sec:open_challenges} discusses challenges and future directions, and Section~\ref{sec:conclusion} concludes the paper.

\section{Preliminaries}~\label{sec:foudations}
Consider a classification problem (supervised learning), where a model $f$ with parameters $\theta$ maps inputs $x \in \mathcal{X}$ to an $n$-class probability distribution, $f_{\theta} : X \rightarrow [0, 1]^n$. The probability assigned to $y$ is denoted as $f(x)_y$. Let $D$ be a training dataset, sampled from some underlying distribution $\mathbb{D}$, where $N$ is the number of data instances. We denote $f_\theta \leftarrow \mathcal{T}(D)$ as a $f$ trained model with parameters $\theta$ that is learned by running the training algorithm $\mathcal{T}$ on the training set $D$.

\textbf{Membership Inference Attacks.}
Successful MIA occurs when an adversary correctly determines if a record belongs to the training set of $f_{\theta}$ or not. In the former case, the record is said to be a member, and in the latter case the record is a non-member. 
We define membership inference inspired in a standard security game~\cite{Yeom,MIA-first-prin,rmia,jayaraman2021revisiting}.

\begin{definition}[Membership Inference Game]
The game proceeds between a challenger and an adversary. Let $\mathbb{D}$ be  the data distribution and $\mathcal{T}$ the training algorithm.
\begin{enumerate}
    \item The challenger samples a training dataset $S \leftarrow \mathbb{D}$ and trains a model $f_\theta \leftarrow \mathcal{T}(S)$ on $S$. 
    \item The challenger flips a bit $b$. if $b=1$, it randomly samples a data point from the training set $(x,y) \leftarrow S$.
    Otherwise, it selects a point from the distribution $(x,y) \leftarrow \mathbb{D}$ such that $(x,y) \notin S$.
    \item The challenger sends the target model $f_\theta$ and target data point $(x,y)$ to the adversary.
    \item With access to the distribution $\mathbb{D}$, the adversary computes $Score_{MIA}((x,y);f_\theta)$, and outputs a membership prediction bit $\hat{b} \leftarrow MIA((x,y);f_\theta)$.
    \item The output of the game is $1$ if $\hat{b} = b$, and $0$ otherwise.
\end{enumerate}
\end{definition}

The prior knowledge of the adversary with respect to the target model, as well as the attack approach, are important factors that can greatly influence the success of MIAs.
Concerning the first factor, the adversary can have white-box, grey-box, or black-box knowledge. 

In a white-box attack, adversaries have complete knowledge about a target model (architecture and parameters), such as gradients and outputs of the hidden layers, which facilitates the membership inference. Although white-box attacks are more common, black-box attacks are more realistic for ML systems as essential model information, such as model implementation details, is often confidential and protected from normal users interacting with the system. 

Black-box attacks assume that the adversary has no access to the target model's internal parameters or training process and can interact with it only through queries. The literature distinguishes three levels of adversarial knowledge: \textit{i)} access only to the predicted label, \textit{ii)} access to confidence scores or probabilities, and \textit{iii)} additional knowledge of the target model family, training procedure, or data distribution. Although the latter is often classified as black-box, it can also be regarded as a grey-box setting due to the availability of auxiliary information beyond query access.

Let us focus on the most predominant setting: score-based attacks relying on the model's output probabilities. Once the training process is completed, the target model $f_{\theta}$ outputs, for each input $x$, a probability vector $p \in \mathbb{R}^k$ of the form $p=(p_1,p_2,\ldots,p_k)$, where $p_i \in [0,1] \forall i$ and \(\sum_{i=1}^{k} p_i=1\). The predicted class label for a feature vector $x$ is the class with the highest probability value in $p$, thus \(\hat{y}=argmax_{i \in \{1,...,k\}}f_\theta(x)_i\).
The output distribution $p = f_\theta(x)$ is often informative about whether a point ($x,y$) was included in the training dataset $D$.
During the attack, a membership score $Score_\mathrm{MIA}((x,y),f_\theta)$ is assigned to each pair. A binary membership decision is then obtained by comparing this score against a threshold $\beta$ which distinguishes members from non-members.

\begin{equation}~\label{eq:mia_attack}
\text{MIA}(x; \theta) = \mathbbm{1}\!\left[\text{Score}_{\text{MIA}}(x; \theta) \geq \beta \right],
\end{equation}

where \(\mathbbm{1}(\cdot)\) is an indicator function that results in 1 if the event occurs, otherwise 0.

\textbf{MIAs in overfitted vs well-generalized target models.}
A successful MIA is typically associated with a high level of overfitting of the target model~\cite{SoK,Salem-ML-Leaks,Shokri-MIA-Against-ML}. Overfitting occurs when a model fits too closely to its training data, resulting in high train accuracy results but fails to make accurate predictions in unseen data. This phenomenon is also associated to memorization, in which larger models are prone to memorizing the training data, thus making them more vulnerable to MIAs~\cite{MIA-first-prin, li2024privacy}. Consider a simple loss function $L(f_\theta(x)_y)$, where $R(\theta)=1/n\sum^{n}_{i=1}L(f_\theta(x)_y)$ is the average loss on the training data and $\hat R(\theta)$ the expected loss on unseen data. If the margin of error between $\hat R(\theta)$ and $R(\theta)$ is large and positive, it is said that the model is overfitting. Nevertheless, it has been shown that although overfitting is sufficient it is not necessary for the success of such attacks~\cite{Yeom,sablayrolles2019white,Stolen-Memories}; well-generalized models can leak membership information, when there are vulnerable data points that cause unique influences on the learning target model, such as outliers~\cite{Long-MIA-Well-Gen}.

\section{Review of Membership Inference Attacks}~\label{sec:sota}
\begin{figure}[!ht]
\centerline{\includegraphics[width=\columnwidth]{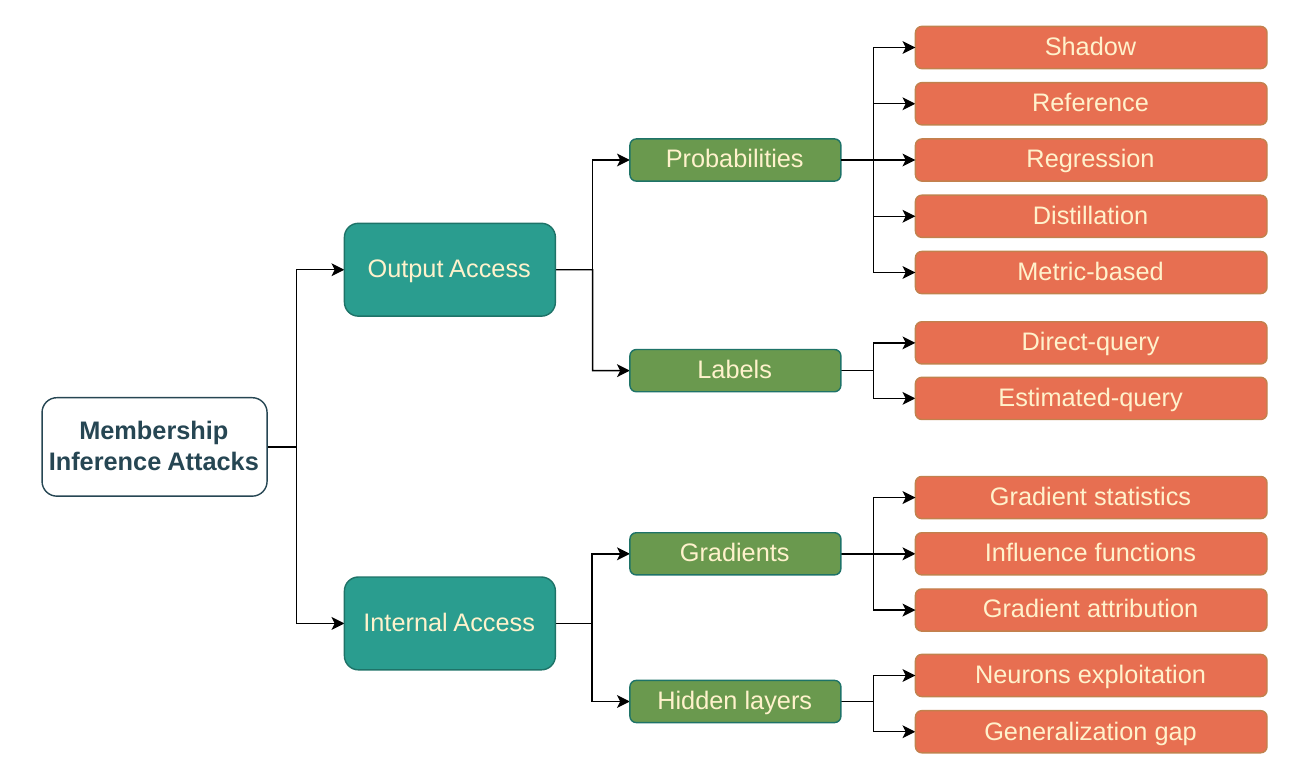}}
\caption{Taxonomy for Membership Inference Attacks.}
\label{fig:mia_taxonomy_tree}
\end{figure}

MIAs are commonly divided w.r.t. the type of adversary's knowledge, namely, black-box or white-box~\cite{SoK,rao2024privacy,niu2025comparing}.
However, a white-box adversary can perform any black-box attack. Therefore, we propose a new categorization based on the information accessible to the adversary. Figure \ref{fig:mia_taxonomy_tree} presents the new taxonomy for MIAs. We divide it into two distinct target model access scenarios. \textit{Output Access}, in which an adversary only has access to predicted labels, confidence scores, or probability vectors, and \textit{Internal Access}, in which an adversary has access to information such as model architecture, parameters, gradients, activations, or training losses. 

Given the vast state-of-the-art in this topic, we provide in Table~\ref{tab:summary} the proposed MIAs. For approaches that were not explicitly named in the original publications, we assign descriptive labels based on their key characteristics to facilitate comparison and discussion.

\begin{table*}[ht!]
\scriptsize
\centering
\resizebox{0.9\textwidth}{!}{%
\begin{tabular}{lllllllll}
\toprule
\textbf{Year} &
\textbf{\begin{tabular}[c]{@{}l@{}}Approach\\ Name\end{tabular}} &
\textbf{\begin{tabular}[c]{@{}l@{}}Attack\\ Classification\end{tabular}} &
\textbf{\begin{tabular}[c]{@{}l@{}} Attack\\ Model-based \end{tabular}} &
\textbf{\begin{tabular}[c]{@{}l@{}}Learning\\ Type\end{tabular}} &
\textbf{\begin{tabular}[c]{@{}l@{}}Adversary\\ Knowledge\end{tabular}} &
\textbf{Defenses} &
\textbf{\begin{tabular}[c]{@{}l@{}}Data\\ Type\end{tabular}} &
\textbf{\begin{tabular}[c]{@{}l@{}}Open\\ Source\end{tabular}} \\
\hline

2017 & MIA~\cite{Shokri-MIA-Against-ML}
& Shadow & \checkmark & DL & \CIRCLE & \checkmark & T \& I & x \\

2018 & LOSS~\cite{Yeom}
& Metric-based & x & DL \& TL & \CIRCLE & x & T \& I & x \\



2019 & MALT~\cite{sablayrolles2019white}
& Metric-based & x & DL \& TL & \CIRCLE & x & I & x \\
2019 & MAST~\cite{sablayrolles2019white}
& Shadow & \checkmark & DL \& TL & \CIRCLE & x & I & x \\
2019 & MATT~\cite{sablayrolles2019white}
& Influence functions & x & TL & \Circle & x & I & x \\

2020 & Prediction uncertainty~\cite{Passive-Active-WB-MIA}
& Gradient statistics & \checkmark & DL & \Circle & x & T \& I & x \\

2020 & Idiosyncratic MIA~\cite{Stolen-Memories}
& Gradient attribution & \checkmark & DL & \Circle & \checkmark & T \& I & x \\

2021 & Merlin \& Morgan~\cite{jayaraman2021revisiting}
& Metric-based & x & DL & \CIRCLE & \checkmark & T \& I & \href{https://github.com/bargavj/EvaluatingDPML}{\faGithub} \\

2021 & Privacy risk score~\cite{Systematic-eval}
& Metric-based & x & DL & \CIRCLE & \checkmark & T \& I & \href{ https://github.com/inspire-group/membership-inference-evaluation}{\faGithub} \\

2021 & Data augmentation attack~\cite{Label-MIA}
& Direct-query & x & DL & \CIRCLE & \checkmark & T \& I & \href{https://github.com/cchoquette/membership-inference}{\faGithub} \\

2021 & Boundary~\cite{li2021label}
& Direct-query & x & DL & \CIRCLE & \checkmark & I & \href{https://github.com/zhenglisec/Label-Only-MIA}{\faGithub} \\


2022 & Attack R~\cite{Enhanced-MIA-Game}
& Reference & \checkmark & DL & \CIRCLE & \checkmark & T \& I & \href{https://github.com/privacytrustlab/ml_privacy_meter/tree/295e7e37e889e12df4083b812f71ed2e2ddd8b4a/research/2022_enhanced_mia}{\faGithub} \\

2022 & Attack D~\cite{Enhanced-MIA-Game}
& Distillation & \checkmark & DL & \CIRCLE & \checkmark & T \& I & \href{https://github.com/privacytrustlab/ml_privacy_meter/tree/295e7e37e889e12df4083b812f71ed2e2ddd8b4a/research/2022_enhanced_mia}{\faGithub} \\

2022 & Attack P~\cite{Enhanced-MIA-Game}
& Population & x & DL & \CIRCLE & \checkmark & T \& I & \href{https://github.com/privacytrustlab/ml_privacy_meter/tree/295e7e37e889e12df4083b812f71ed2e2ddd8b4a/research/2022_enhanced_mia}{\faGithub} \\

2022 & LiRA~\cite{MIA-first-prin}
& Reference & \checkmark & DL & \LEFTcircle & \checkmark & I & \href{https://github.com/tensorflow/privacy/tree/master/research/milira2021}{\faGithub} \\

2022 & Difficulty Calibration~\cite{watson2022importance}
& Reference \& Direct-query & \notcheckmark & DL & \LEFTcircle & \checkmark & T \& I & \href{ https://github.com/facebookresearch/calibration_membership}{\faGithub} \\

2022 & ASTER~\cite{liu2022membership}
& Metric-based & x & DL \& TL & \LEFTcircle & x & T \& I & \href{https://github.com/Wangyiiiiii/Aster}{\faGithub} \\

2022 & SIF~\cite{cohen2024membership}
& Influence functions & x & DL & \Circle & \checkmark & I & \href{https://github.com/giladcohen/sif_mi_attack}{\faGithub} \\

2022 & TrajectoryMIA~\cite{liu2022membership_trajectory}
& Distillation & \checkmark & DL & \CIRCLE & x & I & \href{https://github.com/DennisLiu2022/Membership-Inference-Attacks-by-Exploiting-Loss-Trajectory}{\faGithub} \\

2023 & QMIA~\cite{bertran2023scalable}
& Regression & \checkmark & DL & \CIRCLE & x & T \& I & \href{https://github.com/amazon-science/quantile-mia}{\faGithub} \\

2023 & GapMIA~\cite{White-Box-MIA-DL}
& Generalization gap & x & DL & \Circle & \checkmark & T \& I & x \\

2024 & RMIA~\cite{rmia}
& Reference & \checkmark & DL \& TL & \CIRCLE & x & T \& I & \href{https://github.com/privacytrustlab/ml_privacy_meter/tree/d32734161a3395211fe5f3cd461932290b1fafbe/research/2024_rmia}{\faGithub} \\

2024 & IHA~\cite{suri2024parameters}
& Influence functions & x & DL \& TL & \Circle & x & T \& I & \href{https://github.com/iamgroot42/auditingmi}{\faGithub} \\

2024 & LiRA-L~\cite{suri2024parameters}
& Reference & \checkmark & DL \& TL & \CIRCLE & x & T \& I & \href{https://github.com/iamgroot42/auditingmi}{\faGithub} \\

2024 & GLiRA~\cite{galichin2025glira}
& Distillation & \checkmark & DL & \CIRCLE & x & I & x \\


2024 & RAPID~\cite{he2024difficulty}
& Reference & \checkmark & DL & \CIRCLE & \checkmark & I & \href{https://github.com/T0hsakar1n/RAPID}{\faGithub} \\


2024 & SeqMIA~\cite{li2024seqmia}
& Distillation & \checkmark & DL & \CIRCLE & \checkmark & I & \href{https://github.com/AIPAG/SeqMIA}{\faGithub} \\

2024 & YOQO~\cite{wu2024you}
& Estimated-query & \checkmark & DL & \CIRCLE & \checkmark & T \& I & \href{https://github.com/WU-YU-TONG/YOQO}{\faGithub}  \\

2024 & OSLO~\cite{peng2024oslo}
& Estimated-query & \checkmark & DL & \CIRCLE & \checkmark & I & \href{https://github.com/AIPAG/DHAttack}{\faGithub}  \\

2025 & RaMIA~\cite{tao2025range}
& Reference & \checkmark & DL & \CIRCLE & x & T \& I & \href{https://github.com/taojiashu/ml_privacy_meter/tree/ramia_code/research/2025_ramia}{\faGithub}  \\

2025 & DHAttack~\cite{li2025enhanced}
& Estimated-query & \checkmark & DL & \CIRCLE & \checkmark & T \& I & \href{https://github.com/AIPAG/DHAttack}{\faGithub}  \\

2026 & SHAPOOL~\cite{bai2026toward}
& Reference & \checkmark & DL & \CIRCLE & x & I & \href{https://github.com/BaiLibl/ShadowPool}{\faGithub}  \\

2026 & IMIA~\cite{du2026imia}
& Reference & \checkmark & DL & \CIRCLE & \checkmark & T\&I & \href{https://github.com/zealscott/IMIA}{\faGithub}  \\


\bottomrule
\end{tabular}
}
{\footnotesize
\CIRCLE  represents the black-box setting, \Circle the white-box and \LEFTcircle means that proposed attack is tested on both.
}
\caption{Overview of membership inference attack proposals and their main characteristics. For each attack, we indicate whether it employs auxiliary models (e.g., shadow models, reference models or/and attack classifiers/regressors), the target model type (deep learning (DL) or traditional learning (TL)), the data types considered (tabular (T) and/or image (I)), and whether the attack was evaluated against defense mechanisms. "Open
Source" contains links to repositories.}
\label{tab:summary}
\end{table*}




\subsection{Output Access}

Model outputs include probabilities, logits and predicted labels. Since logits and probabilities reveal similar levels of confidence, we treat them as a single category, distinguishing only between probabilities and labels access.

\textbf{Probabilities.} 
One of the pioneers MIAs in ML~\cite{Shokri-MIA-Against-ML} perform the attack by training multiple \underline{shadow} models that mimic the behavior of the target model. In a black-box setting with access to output probabilities and partial architectural knowledge, multiple shadow models are trained on datasets sampled from the same distribution as the target model. The known membership status of the shadow data is then used to construct an attack dataset of the form $\{\textit{shadow model's top probabilities, ground-truth label, } x\}$, where $x\in\{0,1\}$ indicates whether a sample is a member. The attack is then formulated as a binary classification problem to infer membership. Numerous studies have demonstrated its effectiveness~\cite{MIA-in-MLaaS, li2021membership, niu2025comparing}.

However, this approach presents some challenges. First, the success of the attack depends on the knowledge of the data distribution and target model's architecture. 
Second, it requires multiple shadow models, which is computationally expensive, especially when the model under attack is large. 
Therefore, subsequent works show improvements in MIAs by restricting knowledge assumptions and using limited shadow models -- in some cases, just one~\cite{Yeom, Salem-ML-Leaks}, as we also demonstrate in Section~\ref{sec:reference_models}. 
In addition, MIAs have been shown to be effective to some extent through data transfer attacks~\cite{Salem-ML-Leaks} or model transferability~\cite{MIA-in-MLaaS}.

A major limitation is its poor performance at low false-positive rates, as it relies only on the mean loss values. Consequently, the attack becomes suboptimal when individual samples have markedly different loss scales and variances~\cite{MIA-first-prin}. To address this issue, a principled parametric approach models the loss distributions using well-calibrated Gaussian likelihood estimates.
The attack proceeds by training multiple \underline{reference} models, half containing the target example $(x,y)$ (IN models) and half excluding it (OUT models). Each reference model $f'_{\theta}$ is queried to obtain the correct class logits, $f'_{\theta}(x)y$. Separate Gaussian distributions, $\mathcal{N}(\mu_{IN},\sigma^2_{IN})$ (similar for the OUT models), are then fitted to the IN and OUT logits. Finally, the likelihood ratio between the two distributions is computed and used as the membership score.

Reference models are similar to shadow models but serve a different purpose. While shadow-based attacks mimic the target model, reference-based attacks estimate output likelihoods under the IN and OUT hypotheses, often achieving better performance~\cite{Hierarchical-tree-based, yichuan2024assessing, suri2024parameters,tao2025range,rmia,Enhanced-MIA-Game,du2026imia}.
Furthermore, a simple modification of the score, called difficulty calibration, can drastically improve the reliability of the attack~\cite{watson2022importance,he2024difficulty}. A low loss sample is not necessarily a member; it could also be due to its low difficulty. Difficulty calibration assumes that outputs from reference models can effectively represent the difficulty of target samples, thus improving existing MIAs.

Despite such improvements, previous studies are still computationally costly.
A lightweight alternative uses a mixture of experts architecture~\cite{bai2026toward}. It works by replacing independent shadow models with shared models derived from a single mixture of experts training process. This can reduce costs by approximately 75\% compared to the most applied attack (RMIA~\cite{rmia}) with an equivalent attack efficiency. Nevertheless, it requires a sufficient amount of auxiliary data, which may not be a realistic scenario in many cases.

A different approach leverages \underline{distillation}, where a "teacher" model transfers knowledge to a "student" model. Reference-based attacks have been extended to this setting~\cite{Enhanced-MIA-Game,galichin2025glira,jagielski2023students}. In particular, self-distillation, where both models share the same training data, can amplify overfitting and improve attack performance~\cite{jagielski2023students}. This occurs because memorized samples lead to overly confident predictions, increasing the separation between members and non-members.

Knowledge distillation also serves as a proxy for reconstructing training dynamics by tracking how membership signals evolve across training epochs. This strategy captures this evolution as a loss trajectory~\cite{liu2022membership_trajectory,li2024seqmia}. 
Naturally, distillation-based attacks must train and distil one or more local models over many epochs, which incurs in high computational cost. Moreover, it requires access to an auxiliary dataset, which is rarely identical to the target training data, and therefore, the distribution gap degrades signal quality.

To overcome the computational cost of training multiple shadow or reference models, a model-agnostic approach was proposed to eliminate the need for such models~\cite{bertran2023scalable}. Instead, it trains a quantile \underline{regression} model using only member and non-member samples to perform membership inference.
Nevertheless, the quantile regression quality is sensitive to the volume of the data, i.e. on small datasets, the regression model may not learn accurate per-example conditional quantiles. 

In contrast, \underline{metric-based} approaches perform membership inference directly from statistics derived from the target model outputs, avoiding attack model training and, in many cases, the need for auxiliary data~\cite{MIA-ML-Survey}.
For example, MIA based on the prediction correctness metric requires only knowledge of the predicted labels. The attack follows this principle: if the target model correctly predicts a record, then the adversary infers it as a member, otherwise as a non-member~\cite{Yeom}. 
Several model-free MIAs infer membership by comparing a metric derived from predictions against a threshold. One approach uses the prediction loss, assuming that training members exhibit lower cross-entropy loss than non-members~\cite{Yeom}. Others rely on the maximum prediction probability or prediction entropy~\cite{Salem-ML-Leaks,sablayrolles2019white}. The former exploits the tendency of models to assign higher confidence to training members, whereas the latter leverages the higher uncertainty of non-members, reflected by larger entropy values. However, prediction entropy ignore the ground-truth label, which can lead to misclassification of members and non-members.
The modified prediction entropy metric overcomes this limitation by incorporating the ground-truth label into its computation~\cite{Systematic-eval}. 

Although effective, these scalar metrics only consider the model output at the original input and ignore how predictions behave under perturbations. An extension to loss-based attacks analyzes the variation of loss after small input perturbations. Training members are more likely to lie near local minima due to overfitting, causing perturbations to increase their loss. Further improvement combines the original loss with the perturbation-based score using multiple thresholds~\cite{jayaraman2021revisiting}.

Subsequent work introduced more informative metrics by comparing samples against population distributions~\cite{Enhanced-MIA-Game}. Membership is inferred by comparing the target sample against the output distribution estimated from this population. Unlike reference-based attacks, uncertainty arises only from the queried sample, not from retraining different models.

Finally, a few works exploit the local geometry of the model. In particular, Jacobian matrix-based attacks quantify how sensitive the prediction is to small perturbations of the input, providing an additional membership signal. This matrix reflects the relationship between each feature's perturbation and the corresponding prediction's change~\cite{liu2022membership,central-fl-comp}.
Samples are clustered based on sensitivity, with the lowest-sensitivity cluster identified as members. However, selecting an appropriate $\epsilon$ for Jacobian approximation can be time-consuming.


\textbf{Label.} 
A big concern of the majority MIAs is that the confidence (or loss) scores of the model are exposed, which rely on continuous-valued predictions. For a higher level of protection, prior work has considered obfuscating the model's output by returning only the top label or modifying predicted values for a higher level of protection. Given a target model, the adversary queries an input record and obtains only the predicted label.
Despite this limited information, \underline{direct-query} label attacks can still identify whether a record was in the original training data. The formulation is: if a sample is a member, it tends to be harder to perturb across the decision boundary than a non-member, thus the attack measures how the label changes under perturbations. Membership can be inferred by exploiting data augmentation and decision boundary to evaluate robustness of the target records to these input perturbations~\cite{Label-MIA,li2021label}. 
Besides, these strategies can also benefit from difficulty calibration~\cite{watson2022importance}, making them more effective.

However, direct-query label attacks typically require many queries to estimate prediction robustness, making them computationally expensive and potentially impractical. In contrast, \underline{estimated-query} attacks replace this iterative process with offline estimation, enabling MIAs from a single query to the target model~\cite{peng2024oslo,wu2024you}.
These strategies employ IN/OUT surrogate models during an offline phase to estimate the information required for single query inference. While these approaches need only one query on the target model, it requires crafting a query sample across all IN and OUT models.
Increasing a little the query budget, a shadow-based proposal aims to approximate the non-member state where its boundary distances are then modeled as Gaussian distributions~\cite{li2025enhanced}. Although there is additional computational overhead, this approach presents a higher attack effectiveness.
Similarly to this shadow training, distillation has also been tested for label-based attacks~\cite{liu2022membership_trajectory}. However, it requires much higher query budget. Despite recent progress, label-based attacks are underexplored and do not work well for non-images data due to the method of measuring the boundary distance.

\textit{In summary, attacks that rely on the output access of the target model are the most practical and widely applicable. Probability-based attacks generally offer the best performance by exploiting confidence scores. Nonetheless, label-based attacks remain applicable even when only predicted labels are exposed. Despite this, their effectiveness is limited by the amount of information available through the model outputs, which restricts the membership signal that can be exploited. Yet, they can achieve a favorable trade-off between practicality and effectiveness. 
}

\subsection{Internal Access}~\label{sec:internal_access}

Although previous attacks are carried out with limited knowledge, several researchers have shown how to improve the attack success in deep learning models~\cite{sablayrolles2019white,Passive-Active-WB-MIA,cohen2024membership,White-Box-MIA-DL,Stolen-Memories,suri2024parameters}. They assume the adversary has access to the target model's internals, corresponding to a white-box scenario.

\textbf{Gradients.} One of the first proposals demonstrates how to exploit the privacy vulnerabilities of the stochastic gradient descent (SGD) algorithm~\cite{Passive-Active-WB-MIA}. Since SGD updates the model parameters by minimizing the loss over small subsets of the training set, the per-sample gradients encode information about whether a record was used for training. 
The attack leverages these \underline{gradient statistics} to distinguish members from non-members.


Moreover, exploiting the \underline{influence} of individual samples in the learned parameters can drive the membership signal~\cite{sablayrolles2019white}. The attack measures the alignment between the parameter displacement and the loss gradient to estimate whether a sample contributed to training. However, it has limited applicability to neural networks because the required assumption of a unique global minimum for the Laplace approximation does not hold; it has been evaluated only on logistic regression.
More recent work estimates sample influence through Hessian-weighted self-influence, measuring how strongly a sample affects its own prediction~\cite{cohen2024membership}. By introducing parameter-gradient alignment, this approach bridge the gap between loss-based attacks~\cite{Yeom} and influence-function-based approaches~\cite{sablayrolles2019white}, providing a more principled estimation of sample-specific influence through Hessian-weighted self-influence.
To further improve discrimination, self-influence is combined with an influence alignment term that measures the agreement between the target sample's influence and the average influence of the remaining training data~\cite{suri2024parameters}. This unified score yields stronger membership signals but requires access to the full training set to compute the average influence.

Additionally, some records in the target model's training set may contain specific features that identify them uniquely.
For instance, if a target model is trained to recognise a particular individual among multiple faces and only one contains a pink background, then this feature could serve as evidence of membership. This demonstrates that privacy issues resulting from overfitting do not necessarily manifest in the output behavior, but rather in the way the model uses features. In other words, the idiosyncratic use of features by a model can provide evidence for membership~\cite{Stolen-Memories}. To uncover such memorized features, the attack employs \underline{gradient attribution} techniques that quantify the contribution of individual input features to the model's prediction. A Bayes-optimal threat model is used to derive a set of parameters that profile idiosyncratic features, which are then used to construct a logistic attack model. 

\textbf{Hidden layers.} Other attacks leverage internal layer representations as the membership signal~\cite{White-Box-MIA-DL,WB-Neurons}. One such approach estimates the KL divergence between the hidden-layer feature distributions of training and test samples, treating this quantity as an internal \underline{generalisation gap}~\cite{White-Box-MIA-DL}. This gap is more pronounced at intermediate layers than at the output layer, with early layers showing a sharp increase that gradually narrows deeper in the network. 
But selecting the most influential subset can improve both the accuracy and efficiency of MIA~\cite{WB-Neurons}. Since only a fraction of \underline{neurons} substantially impacts the target classification task, KL divergence and random forest importance scores are combined to identify the most influential neurons across all layers. SHAP values (explainable AI technique) are then used to quantify each feature's contribution within those neurons and capture their directional influence on the model output. Membership is inferred from the disparity in activation patterns between the neuron sets characteristic of training and unseen samples.
However, estimating KL divergence over high-dimensional feature spaces is susceptible to the curse of dimensionality, limiting scalability to deeper or wider architectures.

\textit{In summary, the richer information obtained from internal access often results in higher attack accuracy. 
However, these methods are generally limited to white-box settings and have been developed primarily for deep learning models. 
Moreover, these attacks typically incur substantial computational overhead due to the need to compute gradients, Hessian approximations, or intermediate activations.} 

\section{Defenses to Membership Inference Attacks}~\label{sec:defenses_sota}
Given the wide range of proposed MIAs, there is a growing need to develop mitigation mechanisms that effectively reduce their success.
Defenses against MIAs can be applied during the training phase or the prediction phase~\cite{MIA-in-MLaaS}. They are typically classified as either model hardening or API hardening methods. The former aims to apply techniques that reduce the target model's level of overfitting. The latter aims to limit the information accessible in the prediction vector. 

\subsection{Model Hardening Defenses} This category covers regularization techniques, 
hyper-parameter tuning, differential privacy, model stacking, knowledge distillation, outlier detection, and k-anonymity.

\textbf{Regularization techniques} help to smooth model confidence distributions which may weaken MIAs. For instance, dropout reduces overfitting by reducing variance~\cite{Shokri-MIA-Against-ML,Salem-ML-Leaks,kaya2020effectiveness}. It randomly sets some nodes to zero during the training process. This effectively creates smaller networks that each need to learn to solve the network's task. Although this mitigation is generally efficient, it is limited to neural networks. 
Similarly, $l$2-regularization~\cite{Shokri-MIA-Against-ML,MIA-in-MLaaS,Label-MIA,li2021label} penalizes large weights to improve generalization and reduce the confidence gap between members and non-members, although excessive regularization may degrade model performance. More recently, entropy-based regularization has been proposed to prevent overconfident predictions by combining high-entropy soft labels with an entropy regularizer during training~\cite{chen2024overconfidence}.

Unlike these methods, adversarial regularization aims to reduce prediction loss by optimizing the target model against membership inference. A common approach is Min-Max privacy game in which the target model is trained using an adversarial process~\cite{Adv-Regul,li2021membership}. Specifically, the gain of the MIA is used as a regularizer for the target model that controls the privacy-utility trade-off. 
Related approaches include adversarial training~\cite{MemGuard,li2024privacy} and data augmentation~\cite{cohen2024membership, Stolen-Memories,li2024privacy, li2021membership}. While both introduce additional training examples to improve robustness, adversarial training may inadvertently increase memorization and privacy leakage, whereas data augmentation alone has shown limited effectiveness against MIAs~\cite{li2024privacy}. 

Multi-sample convex combination~\cite{White-Box-MIA-DL} leverages the generalization gap of the internal layers of the target model. It creates synthetic feature representations in hidden layers by linearly mixing the outputs of hidden units from different data points, resulting in new blended feature representations. This discourages the model from having highly specific memorized patterns in the hidden layers, thus obscuring whether a specific data point was part of the training set or not. This was designed to protect models in a white-box scenario.

Label smoothing or early stopping can also act as regularization techniques~\cite{kaya2020effectiveness}. However, label smoothing can inadvertently increase vulnerability to MIAs. By reducing prediction confidence, it can obscure overfitting signals yet amplify membership cues. Conversely, early stopping consistently mitigates MIAs by limiting overfitting: the earlier training is stopped, the lower the attack success.

\textbf{Hyper-parameter tuning} of ML models must be carefully defined. Certain configurations can lead to a higher risk of MIAs. 
Some models, namely tree-based, have already been tested with multiple combinations of hyper-parameters~\cite{Hierarchical-tree-based} to guide in the parameters selection.
In addition, attention should be given to the choice of target model. For example, the Naive Bayes classifier has been shown to be more resilient to these attacks as it usually does not overfit~\cite{MIA-in-MLaaS}.

\textbf{Differential privacy (DP)} is one of the most widely adopted defenses against MIAs~\cite{Label-MIA,Stolen-Memories,MIA-first-prin,watson2022importance,he2024difficulty,li2021label,cohen2024membership}. By injecting calibrated noise during training, typically to gradients in SGD~\cite{song2013stochastic}, DP limits the influence of any individual record on the learned model. However, several studies have highlighted the difficulty of achieving a satisfactory privacy-utility trade-off. In fact, the privacy budget $\epsilon$, which represents the degree of privacy protection, 
when small, it decreases the effectiveness of the adversary, but also reduces the utility of the target model. To mitigate this trade-off, several variants have been proposed, including combining DP with transfer learning in which the knowledge learned from a task is reused~\cite{Label-MIA} or applying DP only to the top layers of a model~\cite{Stolen-Memories}.
Compared with standard regularization techniques such as dropout, DP-based methods generally provide stronger protection~\cite{Stolen-Memories}.

\textbf{Model stacking} is an ensemble learning that combines multiple base ML models to construct a final model~\cite{Salem-ML-Leaks,li2021membership}. The base models obtain the outputs and concatenate them to apply the result to the final model, which predicts the final output. All base models are trained on disjoint sets of data. However, this strategy is easily circumvented in a white-box setting as the base models are available to the adversary~\cite{Stolen-Memories}.

\textbf{Knowledge distillation} is commonly used for model compression but it also protects against MIAs by thwarting the access of the resulting models to the private training data~\cite{shejwalkar2021membership,jagielski2023students,tang2022mitigating,kaya2020effectiveness}. The intuition is that since the student model is not directly optimized over the private set, their membership may be protected. However, this method can only be applied when a public dataset is available for training.

\textbf{Outlier identification} estimates whether a given instance is an outlier. An outlier can be more likely to be a vulnerable target record when it is in the target model's training set, even if the model is already well-generalized~\cite{Long-MIA-Well-Gen}. After identifying the outliers, the data points that do not significantly contribute to the utility of the model are removed.

\textbf{$k$-anonymity} can be applied to the training data before model training, ensuring that each record is indistinguishable from at least $k-1$ others~\cite{MIA-in-MLaaS}. This reduces the influence of individual records and can mitigate membership leakage. Since records with unique attribute combinations are generally more susceptible to re-identification, they are also expected to be more vulnerable to MIAs. However, this relationship has not yet been systematically investigated.


\subsection{API Hardening Defenses} This category aims to modify the model's external outputs, which includes restriction of prediction vector information and adding noise to the prediction vector.

\textbf{Restriction of prediction vector information} can be achieved using several strategies, for example by: \textit{i)} restricting the prediction vector to the top $k$ classes by adding a filter to the last layer of the target model~\cite{MIA-in-MLaaS}; \textit{ii)} decreasing the precision of the prediction vector by rounding the classification probabilities down to $d$ floating-point digits; and, \textit{iii)} increasing the entropy of the prediction vector by adding or modifying the softmax layer and increasing its normalizing temperature to above zero~\cite{Shokri-MIA-Against-ML}.
However, confidence masking alone has been shown to provide limited protection~\cite{Label-MIA}.

An improved strategy works by replacing confidence scores rather than simply reducing their precision or entropy. It uses low-confidence prediction vectors obtained from randomly generated non-member samples to mask the original scores, while preserving the ordering of predicted classes~\cite{chen2024overconfidence}.

\textbf{Adding noise to the prediction vector} means that crafted noise is carefully added to each confidence score vector of the target classifier~\cite{MIA-in-MLaaS,MemGuard}.
Since the adversary may know the defense mechanism, but the defender does not know the attack classifier, it is essential that noise is added randomly; otherwise, the adversary could learn it and adapt their MIA technique. 
However, the distance between the noisy and true prediction vectors should remain minimal to preserve the trade-off.
Although stated as effective in black-box scenarios, this defense does not modify the generalization gap in the hidden layers of the target model, and thus does not protect against adversaries with access to the model's internals~\cite{White-Box-MIA-DL}.

Despite the notable differences in defensive approaches between model and API hardening, some can be combined to enhance output protection. For example, generalization and perturbation can be applied together~\cite{Long-MIA-Well-Gen}. Nevertheless, API hardening defenses are generally ineffective against white-box adversaries, who can access the original model outputs before post-processing~\cite{Stolen-Memories}.


\section{Membership Inference Evaluation}~\label{sec:miaeval}
Regardless of the underlying attack strategy, MIA evaluation is formulated as a binary classification problem, where the objective is to distinguish between members and non-members. 
All attack evaluations are compared to a clueless adversary. This adversary makes random guesses about the membership and is expected to get an accuracy of 0.5.

MIAs were initially evaluated using \textit{accuracy} to capture overall correctness. For a more comprehensive assessment, complementary measures have been used, such as \textit{precision}, which quantifies how many truly members belong to the training set, and \textit{recall} that measures the attack coverage, i.e. the amount of actual members that the adversary successfully detects~\cite{Shokri-MIA-Against-ML, Yeom, MIA-in-MLaaS, Stolen-Memories}. 
Other similar evaluation metrics include \textit{F1-score}~\cite{WB-Neurons,CS-MIA} and \textit{mean average precision} (mAP)~\cite{sablayrolles2019white}.

These evaluation metrics have been applied when the attack dataset is balanced, which means that the portions of members and non-members are similar.
\textit{Membership advantage}, obtained through the difference between \textit{true-positive rate (TPR)} and \textit{false-positive rate (FPR)}~\cite{Yeom} quantifies how much better an adversary is than random guessing at distinguishing members from non-members, in which a difference of zero corresponds to the random guess. Unlike precision/recall, the analysis of TPR/FPR is independent of the prevalence of members in the population, and thus may be more appropriate when this prevalence is unknown.

However, such metrics are average-case and assign equal cost to false positives and false negatives. Thus, \textit{Receiver Operating Characteristic (ROC)} curve~\cite{MIA-first-prin} and its \textit{Area Under the Curve (AUC)}~\cite{Salem-ML-Leaks} were suggested to MIAs assessment since it emphasizes positive predictions over non-membership predictions. It compares the attack's TPR and FPR for all possible decision threshold. Nevertheless, AUC can be misleading because it averages performance over the entire FPR range, including high false-positive rates that are of limited practical relevance. Consequently, recent work recommends reporting the TPR at a fixed low FPR (e.g., 0.1\% or 0.001\%)~\cite{MIA-first-prin}.

Additionally, for attacks requiring multiple trained models (e.g.~\cite{rmia,MIA-first-prin}), the uncertainty of the estimated TPR should be quantified, using Clopper-Pearson confidence intervals, to account for variability arising from finite sample sizes~\cite{tobaben2026impact}.


A concern remains with the previous MIAs evaluation metrics: they mostly focus on positive membership. But detecting non-membership is still relevant. In particular, FPR or \textit{False Alarm Rate (FAR)} shows how non-members are often misclassified as members~\cite{Rezaei2021}. 
It should be noted, however, that achieving reliable MIA with high accuracy and low FAR is difficult. Then a difficulty calibration was proposed to balance accuracy and FPR with the aim of achieving a better separation between member and non-member scores~\cite{watson2022importance}.
Similarly, instead of relying on a fixed confidence threshold of 50\%, a dynamic threshold is set based on a predefined false-positive tolerance \(\alpha\), which is set to the \(\alpha^{th}\)-percentile of confidence scores from a sample. This ensures a controlled FPR level with differing calibration or confidence distributions~\cite{Stolen-Memories}. 

\section{Effectiveness of MIAs on tabular data}~\label{sec:mia_tabular}
In this section, we present the experimental evaluation of MIAs on tabular data. Specifically, we investigate the impact of dataset characteristics on privacy leakage and assess the effectiveness of reference-based attack strategies. To guide our analysis, we address the following research questions:

\begin{enumerate}[label=\textbf{RQ\arabic*:}]
    \item Which characteristics of tabular datasets contribute more to increased susceptibility to MIAs?
    \item How effective are surrogate reference models in improving MIAs over reference-free approaches? 
\end{enumerate}

\textbf{Threat model. }
We consider MIAs against a trained tabular classifier \(f_\theta\). Given a candidate labelled record \((x,y)\), the adversary's goal is to infer whether \((x,y)\) was included in the target model's training set. The attack output is a membership score, where larger values indicate higher confidence that the sample is a member.

We assume a black-box, confidence-score adversary. The adversary can query the target model on candidate records and observe the model confidence assigned to the true label, \(p_\theta(y \mid x)\), but does not access model parameters, gradients, or training internals. The adversary knows the model architecture family and has access to auxiliary data drawn from the same distribution as the target training data. This auxiliary data is used to train or reuse reference/shadow models and, when required, to estimate population-level confidence distributions. Candidate membership labels are used only for evaluation, not by the adversary at test time.

\textbf{Experimental Setup. }
We evaluate our methods on ten tabular classification datasets spanning a diverse range of dataset sizes, feature dimensionalities, and class cardinalities (Table~\ref{tab:data}, Appendix~\ref{app:data}). They cover several data domains as well.
We evaluate six classifiers including Multilayer Perceptron (MLP), Random Forest (RF), Decision Trees (DT), XGBoost (XGB), Logistic Regression (LR), Naive Bayes (NB). 
Details on model architectures are present in Appendix~\ref{app:model_arch}. Although tabular foundation models such as TabPFN~\cite{hollmann_tabpfn_2025} have recently attracted considerable attention due to their strong performance on small- to medium-sized datasets, often outperforming classical ML approaches~\cite{erickson2026tabarena}, recent work suggests that they may be substantially more vulnerable to MIAs. In particular, successful attacks have been demonstrated with only a limited number of reference models, and in some cases without requiring a separately trained attack model~\cite{carvalho2026privacy}. Furthermore, the in-context learning paradigm employed by these models introduces a distinct attack surface. Thus, we restrict our analysis to classical ML models, which still remain widely used in real-world and industry applications.

We evaluate four state-of-the-art MIAs: RMIA~\cite{rmia}, LiRA~\cite{MIA-first-prin}, Attack-P~\cite{Enhanced-MIA-Game} and QMIA~\cite{bertran2023scalable}. We focus on these attacks, as they are the most applied in the relevant literature.
In all cases, the attacks are evaluated on the same audit candidate set and under the same target-model access assumptions, differing only in how they estimate the non-member/reference confidence distribution used to score membership. 

All experiments are repeated across five independent random seeds. Each seed induces a new random permutation of the candidate pool, resulting in different training contexts and audit splits. Reported results correspond to the mean and standard deviation across seeds.  

The attack effectiveness is then reported using the area under the ROC curve (AUC), which provides a threshold-independent measure of separability between members and non-members (aggregate discrimination). Additionally, we report TPR at low FPR corresponding to practical attack scenarios in which the adversary acts only on highly confident membership predictions.

\subsection{Data Factors Affecting MIAs}~\label{sec:data_factors}

As previously stated, MIA is not defined solely by the generalisation gap. As such, we next evaluate data-level characteristics that most influence the attacks.
Therefore, we present in Table~\ref{tab:data-prop-mia} a Spearman correlation between a dataset/target characteristic and the attack's AUC, computed across datasets. 
\textit{Granularity} measures how fine-grained the target label space is.
Class \textit{imbalance} describes the overall unequal distribution of samples across classes. \textit{Majority class} describes how strongly the largest class dominates the dataset, measured by its percentage of all samples. \textit{Rarity} measures how poorly represented the rare/minority classes are. \textit{Target entropy} measures the Shannon entropy of the target-class distribution.
\textit{Mean |Corr| w/ Target} means the average absolute correlation between each input feature and the target variable.
\textit{Features w/ Outliers} is the number or proportion of features containing detected outlier observations.

\begin{table}[ht!]
\scriptsize
\resizebox{\columnwidth}{!}{%
\renewcommand{\arraystretch}{1.1}
\begin{tabular}{c|ccccccc}
\hline
\textbf{Attack} & \textbf{Granularity} & \textbf{Imbalance} & \textbf{\begin{tabular}[c]{@{}c@{}}Majority \\ Class\end{tabular}} & \textbf{Rarity} & \textbf{\begin{tabular}[c]{@{}c@{}}Target \\ Entropy\end{tabular}} & \textbf{\begin{tabular}[c]{@{}c@{}}Mean |Corr| \\ w/ Target\end{tabular}} & \textbf{\begin{tabular}[c]{@{}c@{}}Features \\ w/ Outliers\end{tabular}} \\ \hline
RMIA     & 0.831* & 0.079  & -0.624  & -0.677* & 0.915* & -0.333 & -0.267 \\
LiRA     & 0.774* & -0.055 & -0.758* & -0.360* & 0.927* & -0.491 & -0.377 \\
Attack-P & 0.604  & -0.261 & -0.818* & 0.006   & 0.830* & -0.442 & -0.359 \\
QMIA     & 0.699* & -0.236 & -0.794* & -0.194  & 0.915* & -0.479 & -0.267 \\ \hline
\end{tabular}%
}
\caption{Spearman rank correlation between dataset/class properties and MIA success. * denotes $p < 0.05$.}
\label{tab:data-prop-mia}
\end{table}

The results show that \textit{Target Entropy} is the dominant data factor contributing to MIA's susceptibility. It is the only significant property across all four attacks and has the highest correlation magnitudes. Follows class \textit{Granularity} and \textit{Majority Class} which contribute to the success of three attacks, but the last is consistently negative, meaning that datasets dominated by one class are harder to attack.
Additionally, \textit{Rarity} tends to show increased success for reference-based attacks.

A possible solution to reduce the impact of \textit{Target Entropy} and \textit{Granularity} on the success of the attack is to merge rare fine-grained classes into broader ones if the task allows. Otherwise, class-conditional regularization, such as stronger dropout or weight decay on low-support classes, may mitigate to a certain extent. However, these are specific to gradient-trained parametric models with a differentiable loss, which not all of our models support.
The \textit{Rarity} factor is discussed further in Section~\ref{sec:singleouts}.






\textbf{Resampling. }
Many resampling strategies have been proposed to tackle imbalance and majority class dominance, in which SMOTE~\cite{chawla2002smote} is the most well known. However, previous work shows that SMOTE is non-private~\cite{ganev2026smote}. As such, recent work has incorporated DP into the generation process of SMOTE. We therefore experiment with PP-SMOTE~\cite{pp-smote} and DP-SMOTE~\cite{zhou2025smote}. Table~\ref{tab:pp-smote} presents the results for PP-SMOTE since both methods produced very similar conclusions. We use $\epsilon \in \{0.01, 0.5, 1\}$ and resample all classes but the majority class. Due to space limitation, we show only results for the three most vulnerable models.

\begin{table}[ht!]
    \centering
    \scriptsize
    \resizebox{\columnwidth}{!}{%
\begin{tabular}{c|c|ccc|ccc|ccc}
\hline
\multirow{2}{*}{\textbf{Attack}} & \multicolumn{1}{l|}{\multirow{2}{*}{\textbf{Model}}} & \multicolumn{3}{c|}{\textbf{COMPAS}}      & \multicolumn{3}{c|}{\textbf{Surgery}}     & \multicolumn{3}{c}{\textbf{MIC}}          \\
                                 & \multicolumn{1}{l|}{}                                & \textbf{0.01} & \textbf{0.5} & \textbf{1} & \textbf{0.01} & \textbf{0.5} & \textbf{1} & \textbf{0.01} & \textbf{0.5} & \textbf{1} \\ \hline
\multirow{3}{*}{RMIA}            & MLP                                                  & -0.4        & 7.9       & 0.5     & -8.9        & 6.3       & 11.1    & -4.8        & 39.3      & 23.3    \\
                                 & RF                                                   & -1.3        & -1.4       & -1.6     & -14.6       & 5.6       & 9.9     & 23.6       & 28.0      & 32.7    \\
                                 & LR                                                   & -1.2        & -0.0       & 0.1     & -1.3        & -0.2       & -0.7     & -9.0        & -8.3       & -7.6     \\ \hline
\multirow{3}{*}{LiRA}            & MLP                                                  & 3.2        & 7.7       & 3.9     & 2.4        & 12.4      & 18.6    & -2.9        & 24.9      & 15.2    \\
                                 & RF                                                   & -0.9        & -1.2       & -1.6     & -3.8        & 9.4       & 11.6    & 14.0       & 17.2      & 19.3    \\
                                 & LR                                                   & -1.5        & 0.2       & -0.1     & -0.9        & 0.0       & 0.0     & 0.2        & 1.1       & 1.0     \\ \hline
\multirow{3}{*}{Attack-P}        & MLP                                                  & 5.3        & 14.2      & 7.0     & 8.8        & 22.8      & 27.4    & 9.6        & 53.1      & 28.5    \\
                                 & RF                                                   & -1.2        & -0.9       & -1.0     & 6.3        & 23.7      & 29.0    & 9.3        & 17.3      & 19.0    \\
                                 & LR                                                   & 0.4        & -0.1       & -0.1     & 2.3        & 0.5       & 0.4     & 5.8        & 1.3       & -0.3     \\ \hline
\multirow{3}{*}{QMIA}            & MLP                                                  & 5.1        & 14.0      & 7.1     & 1.8        & 17.1      & 21.9    & 6.6        & 53.9      & 27.0    \\
                                 & RF                                                   & -2.6        & -2.1       & -1.8     & -9.7        & 5.8       & 12.2    & 5.2        & 10.3      & 10.6    \\
                                 & LR                                                   & -0.2        & -0.6       & -0.7     & 2.3        & -0.3       & -0.5     & 4.8        & 0.1       & -1.7     \\ \hline
\end{tabular}%
}

    \caption{Attack AUC change before and after applying PP-SMOTE across three privacy budgets.}
    \label{tab:pp-smote}
\end{table}

In general, the oversampling in these scenarios amplifies the influence of the few original points rather than hiding them. This is especially notable for the MIC dataset, which contains 8 predictive classes, with approximately 85\% of the samples belonging to the majority class. Such a highly skewed class distribution makes balancing particularly challenging. 
Nevertheless, our results suggest that, under certain configurations, privacy-aware synthetic data generation can still offer a degree of protection, highlighting the need for careful tuning and empirical evaluation. Importantly, these findings highlight the urgent need for the continued development of robust resampling strategies against such attacks. The evaluated approaches show that the practical privacy protection offered by DP is not always sufficient in this context. 



\textbf{Model analysis. }
Overfitting is typically the cleanest model-level proxy already proved~\cite{Yeom}, where MIA advantage is theoretically bounded by the generalization gap. 
Our results (Table~\ref{tab:overfitting-vs-mia}, Appendix~\ref{app:targets_performance}) show that while DT is the model that overfits the most, RF and XGB present a greater leakage than their generalization gap suggests. 
This stress the importance of analysis of other factors.

Another important factor is the training paradigm. For instance, RF, DT and XGB are based on tree ensembles that typically memorize via leaf partitioning but get stochastic regularization from bagging/feature subsampling. On the other hand, MLP, gradient-based nets can directly overfit to individual points' losses, which is reflected in our results. 

Finally, the capacity relative to dataset size, i.e., the parameter count vs the number of samples. In our study, we use a fixed architecture for MLP, the parameters do not adapt to $n$. The capacity/$n$ should gradually decrease as the size of the dataset increases. It is expected that MLP becomes less vulnerable to MIA due to this effect. Future work includes such analysis. For tree-based models the size and depth affecting parameters (e.g. \textit{max\_depth}) are searched per dataset, so their effective capacity already partially adapts to $n$. 


\paragraph{\textbf{Insight 1} \textit{Target entropy and class granularity are the most robust, attack-agnostic leakage predictors.}}



\subsection{On the Use of Reference Models}~\label{sec:reference_models}
Beyond identifying which data-level characteristics make MIAs more effective, we next examine how much adversarial effort is required to exploit that vulnerability, specifically, whether costly surrogate reference models are necessary or whether simpler attacks suffice. 
To test practical equivalence, we apply the Two One-Sided Tests (TOST) procedure. Given a pre-specified equivalence margin $\Delta=0.05$, TOST evaluates two one-sided hypotheses simultaneously: $H_1:\mu_{diff} \leq -\Delta$ and $H_2:\mu_{diff} \geq +\Delta$. Rejecting both, i.e., both one-sided p-values falling below $\alpha$, allows the conclusion that the true mean AUC difference lies within ($-\Delta$, $+\Delta$), and is therefore statistically indistinguishable from practically zero. Following standard practice, we report $p\_{TOST} = max(p_1, p_2)$, which requires $p\_{TOST} < \alpha = 0.05$ for an equivalence claim.

\begin{figure*}[!ht]
    \centering
    \includegraphics[width=0.85\textwidth]{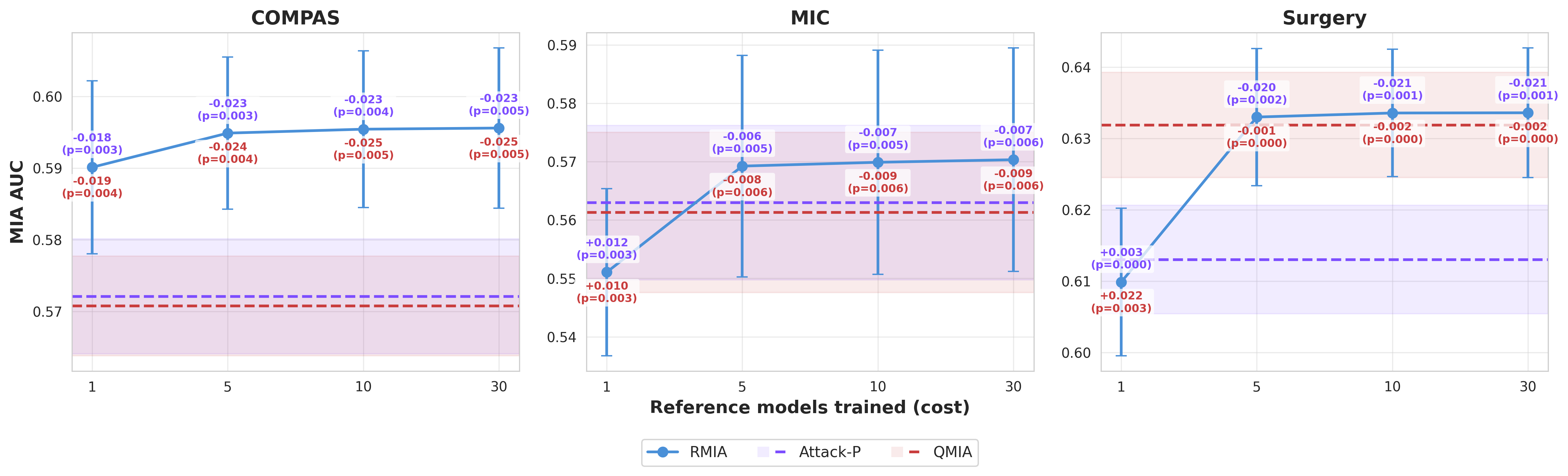}
    \caption{Comparison of Attack-P/QMIA versus surrogate RMIA performance at each reference model budget. The RMIA AUC line represents the mean ± standard deviation across all reference models architectures. The dashed lines show the mean Attack-P and QMIA AUC, with the respective shaded bands depicting the ± Standard Error of the Mean across seeds.}
    \label{fig:proxy-ref}
\end{figure*}

Figure \ref{fig:proxy-ref} applies TOST to the surrogate RMIA, instantiated with reference models drawn from five architectures with MLP fixed as the target model. This reflects a more realistic black-box scenario. For demonstration purposes, we focus solely on one target model, specifically the most vulnerable. We compare all architectures against the two reference-free baselines, Attack-P and QMIA, across three datasets. 

Under the margin $\Delta$, every gap across all three datasets and all reference model budgets falls within ($-\Delta$, $+\Delta$). Even the COMPAS's AUC gap (0.032) which lead for surrogate RMIA, remains inside the equivalence bound. Both one-sided tests reject non-equivalence for every dataset and reference-model budget pair, formally confirming that surrogate RMIA and the reference-free baselines are statistically practical equivalent. Appendix~\ref{app:ref_models} extends this analysis for all datasets. 

Additionally, surrogate RMIA shows consistent diminishing returns in which nearly all AUC gain from increasing the reference model budget is realized by the fifth trained model, with budgets of 10 and 30 producing no further meaningful movement in any dataset, $\Delta \leq 0.002$ relative to the cost of 5 models. This suggests that any advantage from additional surrogate reference models saturates quickly rather than scaling with attacker compute.




\paragraph{\textbf{Insight 2} \textit{Reference-free MIAs are a cost-effective substitute for surrogate reference models, reaching statistically equivalent detection power.}}


\section{Vulnerability of Single-Outs to MIAs}~\label{sec:singleouts}

In this section, we investigate the privacy risks faced by single-out records under MIAs. This record-level analysis deserves a dedicated section, given its singularity. 
We first evaluate whether records identified as single-outs exhibit increased susceptibility to membership inference compared to the general population. Subsequently, we assess the effectiveness of privacy-preserving defenses in mitigating the leakage associated with these high-risk individuals. To guide our analysis, we address the following research questions:


\begin{enumerate}[label=\textbf{RQ\arabic*:}, resume]
\item To what extent are single-out records more vulnerable to MIAs than non-single-out records?
\item How effective are existing defense mechanisms in protecting single-out records against MIAs?
\end{enumerate}

The following results concern seven datasets as they include quasi-identifiers that are easily identifiable.
The set of such attributes is listed in Table~\ref{tab:data_qis} (Appendix~\ref{app:data}). Since singling out is defined with respect to training records, we evaluate susceptibility only among members. All single outs are flagged according to the definition of $k$-anonymity~\cite{sweeney2002k}. Although $k=1$ is the standard setting, as it corresponds to a 100\% re-identification risk, we also consider $k=2$, which still implies a 50\% risk. Importantly, these single-out records are not used to evaluate re-identification; rather, they define groups of records with different levels of uniqueness, which we then use to assess their susceptibility to MIAs. All experimental settings remain unchanged from the previous section.

\subsection{Experimental Results}~\label{sec:singleouts_exp}
For each dataset, model, and attack, we evaluate the difference in attack performance between SO and non-SO members. For TPR-based evaluation, we first calibrate the attack threshold on non-member records to achieve a fixed FPR, and then compare the TPR of SO and non-SO members at that same threshold. This isolates whether being more identifiable in the underlying data translates into increased MIA vulnerability.

\begin{figure}[!ht]
    \centering
    \includegraphics[width=\columnwidth]{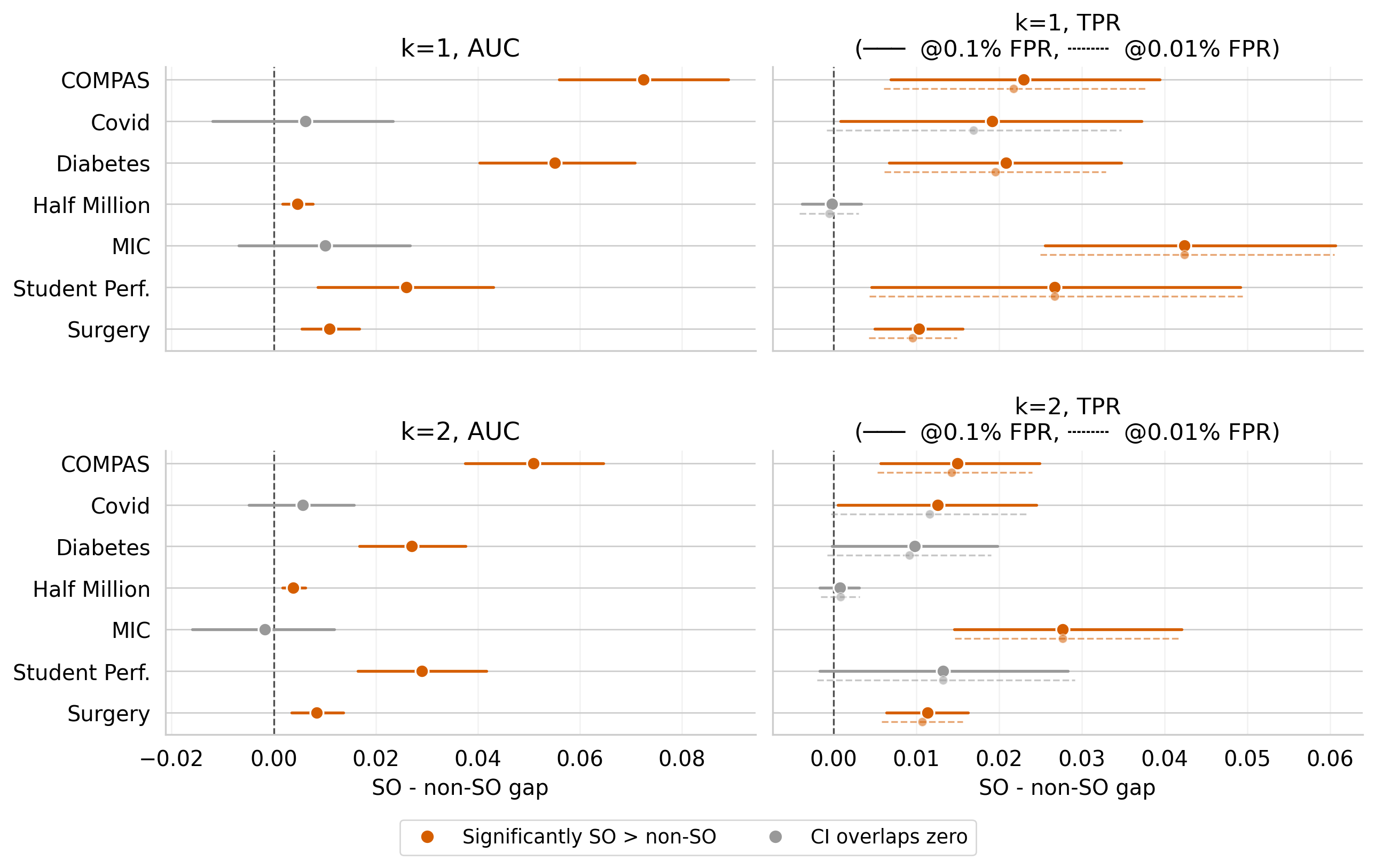}
    \caption{Difference in membership inference vulnerability between singled-out (SO) and non-SO records per dataset, measured by AUC and both TPR@0.1\%FPR (solid) and TPR@0.01\%FPR (dashed) for anonymity $k=1$ and $k=2$. Points show the estimated SO--non-SO difference, with horizontal bars denoting confidence intervals (CIs).}
    \label{fig:singleoutsMIA}
\end{figure}

We quantify this difference using the SO susceptibility gap:
\[
\Delta_{\mathrm{SO}} =
\mathrm{TPR}_{\mathrm{SO}} -
\mathrm{TPR}_{\mathrm{non-SO}}.
\]
Positive values indicate a greater vulnerability among SO records. 
Figure~\ref{fig:singleoutsMIA} shows the SO--non-SO performance gap for two TPR@low FPR at $k=1$ and $k=2$, also including AUC. Each point represents the mean gap pooled across models and attacks, with 95\% bootstrap confidence intervals (CI) over model--seed replicates. When CI overlaps zero indicates that the SO effect is not statistically significant at the level of 5\%.


The SO vulnerability effect is robust across $k$ and metric. The effect is stronger at $k=1$ than at $k=2$, with the gap decreasing significantly as $k$ is relaxed. This matches the intuition where $k=1$ defines a stricter, smaller equivalence class than $k=2$, thus SO records are more extreme records and therefore more susceptible to membership inference, whereas relaxing $k$ dilutes this distinction.
The effect weakens as $k$ increases for both AUC and TPR, with the TPR gap reducing by a comparable, albeit slightly larger, proportion.
At $k=1$, TPR@0.1\% FPR reaches significance in more datasets than AUC (6/7 vs 5/7), while at $k=2$ this reverses (4/7 vs 5/7), as Diabetes and Student Performance lose significance for TPR but not AUC. TPR@0.1\%FPR at $k=1$ shows a strong, robust signal, comparable in strength to looser FPR thresholds (0.5–1\% FPR gave marginally higher pooled effect sizes) and weakening only at the strictest 0.01\% FPR (5/7 datasets significant at $k=1$, dropping to 3/7 at $k=2$). Half Million remains non-significant for TPR across every FPR threshold we tested (0.01\%–1\%) and both values of $k$, despite showing a small but statistically detectable AUC gap. 

Additionally, we observe differences in the strength of the SO-vulnerability effect across attacks in Figure~\ref{fig:singleoutsMIA_attacks} (Appendix~\ref{app:singleouts}). RMIA and LiRA show the clearest and most consistent separation, while Attack-P exhibits a weaker effect at $k=1$ that becomes more pronounced at $k=2$. QMIA shows the weakest evidence for a SO-vulnerability gap.


Overall, singling out increases the adversary's edge over chance by 24.9\% at $k=1$ (95\% CI [16.1\%, 35.2\%]) and 16.5\% at $k=2$ (95\% CI [10.2\%, 24.1\%]).

\paragraph{\textbf{Insight 3:} SO records tend to exhibit greater membership inference vulnerability than non-SO records. This effect is stronger even at the lowest FPR.}

\subsection{Impact of defenses}~\label{subsec:defenses_singleouts}

To understand the effectiveness of defense mechanisms in reducing single-outs' vulnerability to MIAs, we apply two state-of-the-art defenses: DP~\cite{dwork2006differential}~\cite{diffprivlib} and HAMP~\cite{chen2024overconfidence}. The reason these two defenses are chosen is to include both training-time and inference-time defenses. 
Besides, we do not include methods such as MemGuard, based on adversarial examples, as it has demonstrated similar results to the undefended model~\cite{peng2024oslo, chen2024overconfidence}. In addition, regularization defenses, such as early stopping, l2, and dropout, were also discarded since they are model dependent and less efficient than DP~\cite{Stolen-Memories}. 

\begin{figure*}[ht!]
    \centering
    \includegraphics[width=0.85\textwidth]{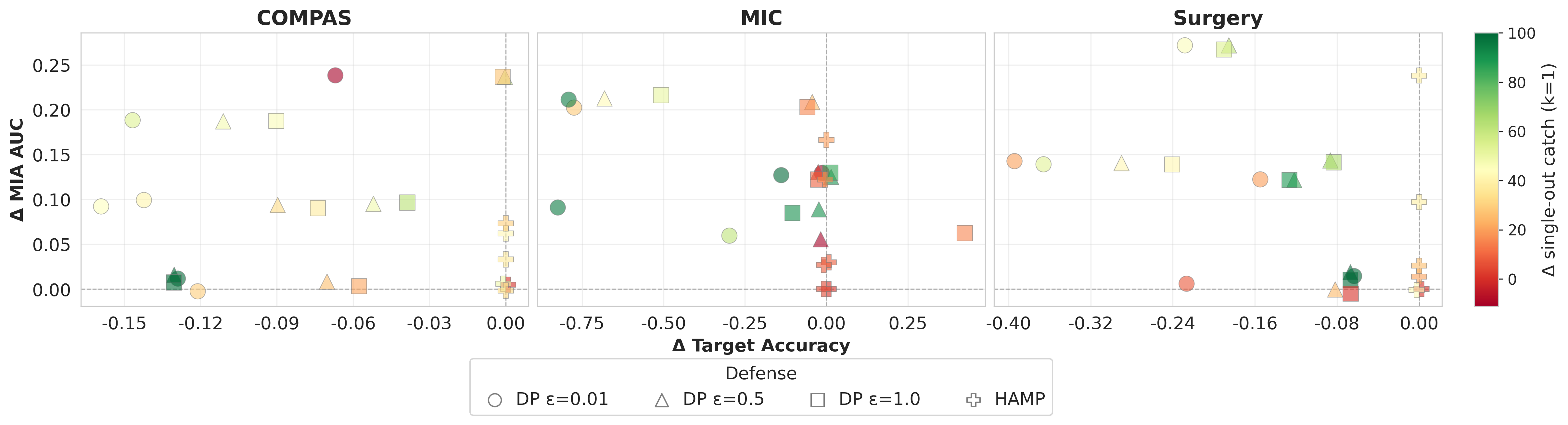}
    \caption{Privacy–utility trade-off of defenses on singled-out ($k=1$) records, by dataset, considering RMIA. Each point is one model. Color shows how much less often single-outs are correctly flagged as members after defense. }
    \label{fig:singleouts-3ds-def}
\end{figure*}

Figure~\ref{fig:singleouts-3ds-def} reports defense privacy-utility trade-off by dataset for $k=1$, considering the baseline as the attack without any defense. 
The plots demonstrate that the attack AUC change is not directly proportional to that of single-out catch.


As evident, HAMP clusters near zero. It costs no accuracy but also buys almost no AUC reduction. The unchanged target model's utility is because this attack is applied during inference time, operating only at the output of the model. Consequently, the reduction in single-out catch is usually small, especially for MIC.
In contrast, DP defenses show a greater reduction in the target model's utility, especially for smaller $\epsilon$ values. 
Despite this, we can find that some DP cases achieve a larger drop in single-out catch than HAMP, while still preserving utility, this is emphasized on MIC and Surgery.
Notably, DP $\epsilon=0.01$ is the costliest but not proportionally the most protective. On MIC in particular, accuracy is reduced to -0.6 and to -0.8 (near-destroying the model) while AUC tops out around 0.2. 
Table~\ref{fig:singleouts-defenses-k1}, Appendix~\ref{app:defenses} extends the analysis to the remaining attacks, which shows that reference-free attacks can capture higher number of single-outs.

Given such results, it is important to apply specific measures to tackle these vulnerable records.

\begin{figure}[ht!]
    \centering
    \includegraphics[width=\columnwidth]{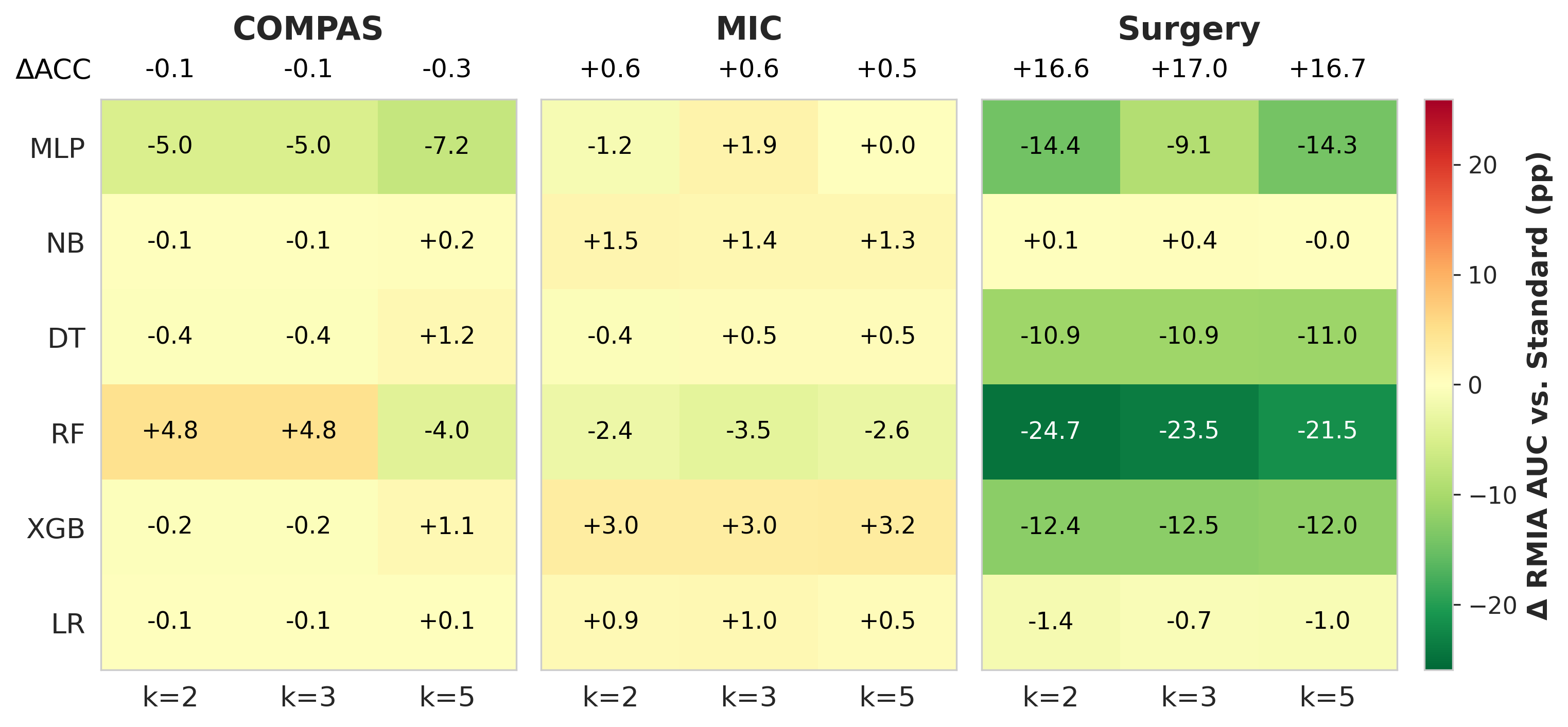}
    \caption{k-anonymity application over datasets and its percentage points difference against the baseline.}
    \label{fig:kanon-3ds}
\end{figure}

\textbf{$K$-anonymous datasets. } We then report in
Figure~\ref{fig:kanon-3ds} the application of $k$-anonymity. We use $k \in \{2,3,5\}$, and both generalization and suppression to the same set of quasi-identifiers.

Overall, we observe a reduction of MIA AUC. Although this is strongly evident for Surgery, COMPAS and MIC show that anonymization can, for certain settings, increase attack success (up to +4.8 pp). Its quasi-identifiers present low-signal attributes, thus applying anonymization methods barely changes the attack's signal. Besides reducing AUC, it also can provide better predictive performance.
These results correlate with the Rarity property mentioned in Section~\ref{sec:data_factors} since rare records are easier to attack because the model memorizes them. By applying k-anonymity's suppression, any record whose quasi-identifier values occur fewer than $k$ times gets those values replaced with a shared placeholder, weakening the attack when those quasi-identifiers are tied to the label.
Future work remains to study the impact of $k$-anonymity on different sets of quasi-identifiers and test on other attacks.

\paragraph{\textbf{Insight 4} \textit{State-of-the-art defenses can lower average vulnerability but singled-out individuals still amplifies MIA.}}


\section{Open Challenges and Research Directions}~\label{sec:open_challenges}
\textbf{Privacy-aware data pre-processing.} Our findings indicate that privacy should be considered before model training rather than only during or after deployment. It is therefore important to extend the evaluation of privacy-aware preprocessing strategies across a broader range of datasets and application domains to better understand their impact on the privacy-utility trade-off. Developing such pipelines may provide a first line of defense against MIAs.

\textbf{Beyond supervised classification.} While classification is the dominant benchmark for MIAs, important learning paradigms remain underexplored, particularly regression models. In regression settings, membership signals are often weaker and harder to calibrate. As models typically exhibit lower overfitting, may have residuals that look similar for members and non-members, making the membership signal noisier and harder to detect~\cite{gupta2021membership}. 
Future research should also investigate privacy risks in other emerging scenarios, including unsupervised learning, tabular foundation models, generative models for synthetic tabular data, and large language models adapted for tabular applications. These settings may exhibit different memorization behavior and require dedicated attack and defense mechanisms.



\textbf{Enhanced attacks.} Current MIAs generally assume that the adversary can query the target model using complete input records. However, this assumption may not be applicable in practice, where only partial information about an individual may be available. Recent studies demonstrate that membership can still be inferred from incomplete feature vectors or feature regions by exploiting the model's latent representations and memorization patterns~\cite{tao2025range,wang2025membershipinferenceattackpartial}. These findings motivate the development of MIAs under more realistic threat models, where adversaries operate with limited or imperfect knowledge of the target records. 


\section{Conclusion}~\label{sec:conclusion}
In this paper, we present a systematization of knowledge for tabular MIAs. We suggest a new taxonomy around the adversary's minimum required access and use it to review a substantial set of attacks and their respective defenses. 
Our results reveal dataset-level properties such as target entropy, granularity, and majority class are strong predictors of a model's overall vulnerability. Most importantly, we show that singled-out records can remain disproportionately exposed even when defenses appear effective, and that $k$-anonymity can amplify rather than reduce this exposure, highlighting the need for record-level and worst-case privacy auditing. Likewise, our evidence that reference-free attacks can match reference-based ones challenges the assumption that stronger adversarial capabilities are necessary for meaningful auditing, suggesting that privacy evaluation can be made substantially cheaper without sacrificing effectiveness.


\section*{Ethical Considerations}
\textbf{Stakeholder-Based Analysis.} We identify three key stakeholders impacted by this research:
\begin{itemize}
    \item \textit{Machine Learning Practitioners.} This work equips ML practitioners training models on tabular data with a systematic understanding of membership inference risk, including which data properties (e.g., singling-out records, class granularity) drive vulnerability, and which common defenses fail to adequately protect against it. This enables more informed decisions about data handling and defense deployment prior to release.
    \item \textit{Data Subjects.} MIAs risk is heightened for singled-out records and rare classes, which we show are disproportionately vulnerable. We mitigate this risk by conducting all experiments exclusively on public benchmark tabular datasets, without targeting real individuals or proprietary models. By identifying which records and defenses are insufficiently protective, our work ultimately helps practitioners to better safeguard data subjects going forward.
    \item \textit{Adversaries.} We acknowledge that our findings, could potentially be misused by adversaries to more effectively target real-world models. We take this risk seriously and discuss concrete mitigations below.
\end{itemize}

\textbf{Mitigations.} We take the following concrete steps to mitigate potential harms:
\begin{itemize}
    \item \textit{Exclusive Use of Public Benchmark Data.} All experiments are conducted exclusively on public tabular datasets, used in accordance with their licenses. We do not target specific individuals or production systems at any point, ensuring that no one's real privacy is compromised in the course of this research.
    \item \textit{Systematic, Reproducible Evaluation.} We conduct a systematic evaluation of established MIAs using standard, well-documented methodology. This transparency allows practitioners and the defense research community to reproduce, verify, and build directly on our findings.
    \item \textit{Preprocessing Defense.} We suggest and evaluate class imbalance and $k$-anonymization as preprocessing strategies to reduce membership leakage. By releasing findings alongside a practical and deployable mitigation, we reduce the risk introduced by this work and provide practitioners with actionable guidance to protect their data.
\end{itemize}

\textbf{Justification for Research and Publication.} Prior work has established that classical MIAs can be effective against tabular models trained on real-world data. However, little is known about which records are most vulnerable, why existing defenses may fail to protect them, and which data properties predict this vulnerability. Our study addresses these gaps, helping practitioners identify privacy vulnerabilities and strengthen protections for vulnerable records, particularly those that are singled out or rare. We further contribute a classification of the MIA literature and a practical tool table summarizing available attacks and their requirements, lowering the barrier for practitioners to audit their own models and adopt these techniques for defensive. 
We emphasize that these findings should support privacy research and responsible deployment, not the targeting of individuals.


\section*{Open Science}
Consistent with open science principles and to support transparency, reproducibility, and further research, we make all the artifacts publicly available at \url{https://anonymous.4open.science/r/sokMIA-4439}. The \texttt{README.md} file provides a detailed, experiment-by-experiment mapping between reported results and the corresponding code required to reproduce them, including environment setup instructions. Each attack is implemented as a standalone script at the repository root, with the respective \texttt{run\_*\_batch.py} files to run for all datasets and models at once.


\bibliographystyle{plainurl}
\bibliography{bib}

\appendix

\section{Experimental Details}\label{app:experiments}
This section describes the full experimental setup of the implemented MIAs.
All experiments were conducted on Ubuntu 24.04 LTS. The system is an NVIDIA GB10 (Grace Blackwell) ARM-based unified-memory node, with a 20-core SoC (10 Cortex-X925 + 10 Cortex-A725) and around 121 GiB of unified CPU/GPU memory shared with a single onboard NVIDIA GB10 GPU. The experiments were run with CUDA 13.0. This setup was used for model inference and attack execution.

\subsection{Data}\label{app:data}

The characteristics of the nine datasets used to evaluate the MIAs are presented in Table~\ref{tab:data}. We use datasets that span a range of sample sizes, features, and labels. The domains include crime, healthcare, finance, education, locations and purchases. 
Table~\ref{tab:data_qis} presents the quasi-identifiers of seven datasets alongside the range of the number of single-outs across the five seeds.

\begin{table}[ht!]
\scriptsize
\resizebox{\columnwidth}{!}{%
\begin{tabular}{l|lcccc}
\hline
\textbf{Dataset}                                                                                                     & \textbf{Domain} & \textit{\textbf{n}} & \textit{\textbf{d}} & \textit{\textbf{c}} &\textit{ \textbf{r}}                                   \\ \hline
COMPAS Recidivism~\cite{ds:compas-recidivism}                                                  & Crime           & 4937             & 20                   & 2         &   1.05                                      \\
Covid~\cite{carvalho2025empowering}                                                            & Healthcare      & 2671             & 29                   & 3         &   6.3                                                                                     \\
\begin{tabular}[c]{@{}l@{}}Diabetes Readmission~\cite{ds:diabetes-readmission}\end{tabular} & Healthcare      & 81410            & 150                  & 2         &    1.17      \\
Half Million~\cite{ds:half_million}                                                           & Finance         & 15855            & 28                   & 12          &    1.08                                    \\
Locations~\cite{ds:Shokri_Datasets,ds:location-unprocessed}                                   & Locations       & 5010             & 446                  & 10         &    3.18                                    \\
Mammography~\cite{ds:mammography} & Healthcare & 11183 & 24 & 2 & 42.01 \\ 
MIC (Myocardial infarction)~\cite{ds:mic}                                                                             &       Healthcare           & 1699             & 122                  & 8          &   119  \\
Purchases-10~\cite{ds:Shokri_Datasets,ds:purchases-unprocessed}                               & Purchases       & 19766            & 600                  & 10       &    5.28            \\
\begin{tabular}[c]{@{}l@{}}Student Performance~\cite{ds:student_performance}\end{tabular}   & Education       & 648              & 30                   & 3         &   3.55                                                  \\
Surgery~\cite{ds:surgery}                                                                      & Healthcare      & 14635            & 24                   & 2                  &    2.97                                                                                                       \\ \hline
\end{tabular}%
}
\caption{Specifications of the evaluation datasets. \textit{n} denotes the number of samples, \textit{d} the number of features, \textit{c} the number of classes, \textit{r} the imbalance ratio.}
\label{tab:data}
\end{table}

\begin{table}[ht!]
\scriptsize
\resizebox{\columnwidth}{!}{%
\begin{tabular}{l|lcc}
\hline
\multicolumn{1}{c|}{\multirow{2}{*}{\textbf{Dataset}}} & \multicolumn{1}{c}{\multirow{2}{*}{\textbf{Selected quasi-identifiers}}}                                                                                                                                               & \multicolumn{2}{c}{\textbf{\# Single-outs}}  \\ \cline{3-4} 
\multicolumn{1}{c|}{}                                  & \multicolumn{1}{c}{}                                                                                                                                                                                                   & \textbf{$k=1$} & \textbf{$k=2$}              \\ \hline
COMPAS                                                 & Age, Race:Native\_American, Race:Other                                                                                                                                                                                 & 16-23          & 34-51                       \\ \hline
Covid                                                  & Age, Cerebro Vascular, Diabetes                                                                                                                                                                                        & 5-10           & 16-25                       \\ \hline
Diabetes                                               & \begin{tabular}[c]{@{}l@{}}time\_in\_hospital, race:Other, age:\{{[}\}50-70), \\ admission\_type\_id:Trauma Center, \\ discharge\_disposition\_id:Discharged to Home, \\ discharge\_disposition\_id:Other\end{tabular} & 11-18          & 23-43                       \\ \hline
Half Million                                           & Gender, State, Country                                                                                                                                                                                                 & 180-216        & 592-620                     \\ \hline
MIC                                                    & Age, Obesity, Chronic Pneumonia                                                                                                                                                                                        & 14-19          & 24-31                       \\ \hline
Student Perf.                                          & School, Address, MJob, School Sup                                                                                                                                                                                      & 4-12           & 13-22                       \\ \hline
Surgery                                                & gender, mort30, mortality\_rsi                                                                                                                                                                                         & 267-313        & \multicolumn{1}{l}{628-669} \\ \hline
\end{tabular}%
}

\caption{Selected quasi-identifiers for each dataset and corresponding number of single-outs range across the 5 seeds.}
\label{tab:data_qis}
\end{table}

\subsection{Model architectures}~\label{app:model_arch}
All models are tuned with Optuna (30 trials, 3-fold CV) to reflect realistic model development practices and obtain competitive configurations. 
The MLP includes three hidden layers (256, 256, 128 units), each followed by a linear Batch Normalization layer and a ReLU activation layer, ending with a linear classification layer. It uses an SGD optimizer with learning rate 0.01, weight decay $10^{-4}$, batch size 256 and 100 epochs. 
We use learning rate over $[10^{-4},10^{-1}]$, weight decay over
$[10^{-6},10^{-2}]$, batch size over $\{128,256,512\}$, and the optimizer
over $\{\mathrm{SGD},\mathrm{Adam},\mathrm{AdamW}\}$.

For RF, Optuna searches over $n\_{estimators}\in \{50,500\}$, $max\_{depth}\in\{None,3,5,8,10,15,20\}$,
${min\_samples\_split}\in\{2,20\}$, $min\_samples\_leaf\in\{1, \max(10,\lfloor n/200\rfloor)\}$ and $max\_{features}$ over a dataset-dependent continuous interval.
In the DT model, the search spans over $max\_depth \in \{4,6,10\}$ and $min\_samples\_leaf \in \{1, 3, 5, 7\}$. 
Moreover, for XGBoost, the $gamma \in \{0, 0.5, 1\}$, $max\_depth \in 	\{5, 10, 15, 20\}$, $learning\_rate \in \{0.1, 0.01\}$, $min\_child\_weight \in \{1, 3, 5, 7\}$, $colsample\_bytree$ and $subsample \in \{0.8, 1\}$ parameters are optimized.
With NB, the adjustment is made to the $var\_{smoothing} \in \{10^k: k \in linspace(-9, 0, 100)\}$ parameter. Finally, LR searches over $C \in \{0.0001, 0.001, 0.01, 0.1, 1, 10, 100\}$, $solver\in \{lbfgs, $\textit{newton-cg}$, saga, sag, $\textit{newton-cholesky}$\}$ and $penalty$ set to $l2$.








\section{Performance of the target models}~\label{app:targets_performance}
Table~\ref{tab:overfitting-vs-mia} shows that DT is the model that overfits the most overall, while LR and NB are the safest models. RF and XGB present a greater leakage than their generalization gap suggests. Meanwhile, MLP's generalization gap and MIA AUC generally track together.
Even though the tree-based models leak more they also show the highest predictive power in average (from 73.1\% to 81.6\%), similarly to MLP's utility with 73.9\% on average.
Notably, LR's lower leakage pairs with its high predictive power (76.4\% average target accuracy), which is just 2.8 pp above the average performance.
NB shows to be the model with the lowest predictive power, with only 55.4\% accuracy (-18.1 pp lower than the average).


\begin{table}[ht!]
\scriptsize
\resizebox{\columnwidth}{!}{%
\renewcommand{\arraystretch}{1.1}
\begin{tabular}{l|cc|cc|cc}
\hline
\multirow{2}{*}{\textbf{Model}} & \multicolumn{2}{c|}{\textbf{COMPAS}}           & \multicolumn{2}{c|}{\textbf{MIC}}              & \multicolumn{2}{c}{\textbf{Surgery}}           \\ \cline{2-7} 
                                & \textbf{Gen. gap (\%)} & \textbf{MIA AUC (\%)} & \textbf{Gen. gap (\%)} & \textbf{MIA AUC (\%)} & \textbf{Gen. gap (\%)} & \textbf{MIA AUC (\%)} \\ \hline
RF                              & 3.1 $\pm$ 2.9          & 59.0 $\pm$ 4.2        & 1.3 $\pm$ 1.1          & 70.6 $\pm$ 5.2        & 6.7 $\pm$ 0.7          & 77.0 $\pm$ 2.5        \\
MLP                             & 10.5 $\pm$ 8.8         & 60.0 $\pm$ 5.8        & 7.0 $\pm$ 4.9          & 59.1 $\pm$ 4.0        & 17.7 $\pm$ 7.9         & 64.3 $\pm$ 5.3        \\
NB                              & 0.8 $\pm$ 1.7          & 50.5 $\pm$ 0.8        & 7.9 $\pm$ 2.0          & 56.0 $\pm$ 1.4        & 0.1 $\pm$ 0.7          & 50.2 $\pm$ 0.7        \\
DT                              & 42.3 $\pm$ 1.2         & 74.2 $\pm$ 0.6        & 18.0 $\pm$ 1.8         & 62.8 $\pm$ 1.2        & 18.9 $\pm$ 0.4         & 62.3 $\pm$ 0.4        \\
XGB                             & 16.3 $\pm$ 1.3         & 68.3 $\pm$ 0.9        & 10.6 $\pm$ 0.5         & 70.9 $\pm$ 2.2        & 6.2 $\pm$ 0.3          & 64.3 $\pm$ 0.7        \\
LR                              & 0.6 $\pm$ 1.3          & 51.5 $\pm$ 0.7        & 10.8 $\pm$ 2.3         & 62.6 $\pm$ 1.2        & 0.2 $\pm$ 0.6          & 51.5 $\pm$ 0.5        \\ \hline
\end{tabular}%
}

\caption{Comparison of the generalization gap of the target model with the RMIA AUC.}
\label{tab:overfitting-vs-mia}
\end{table}


\section{Attack performance on ten datasets}~\label{app:attack_performance}

\begin{figure}[ht!]
    \centering
    \includegraphics[width=\columnwidth]{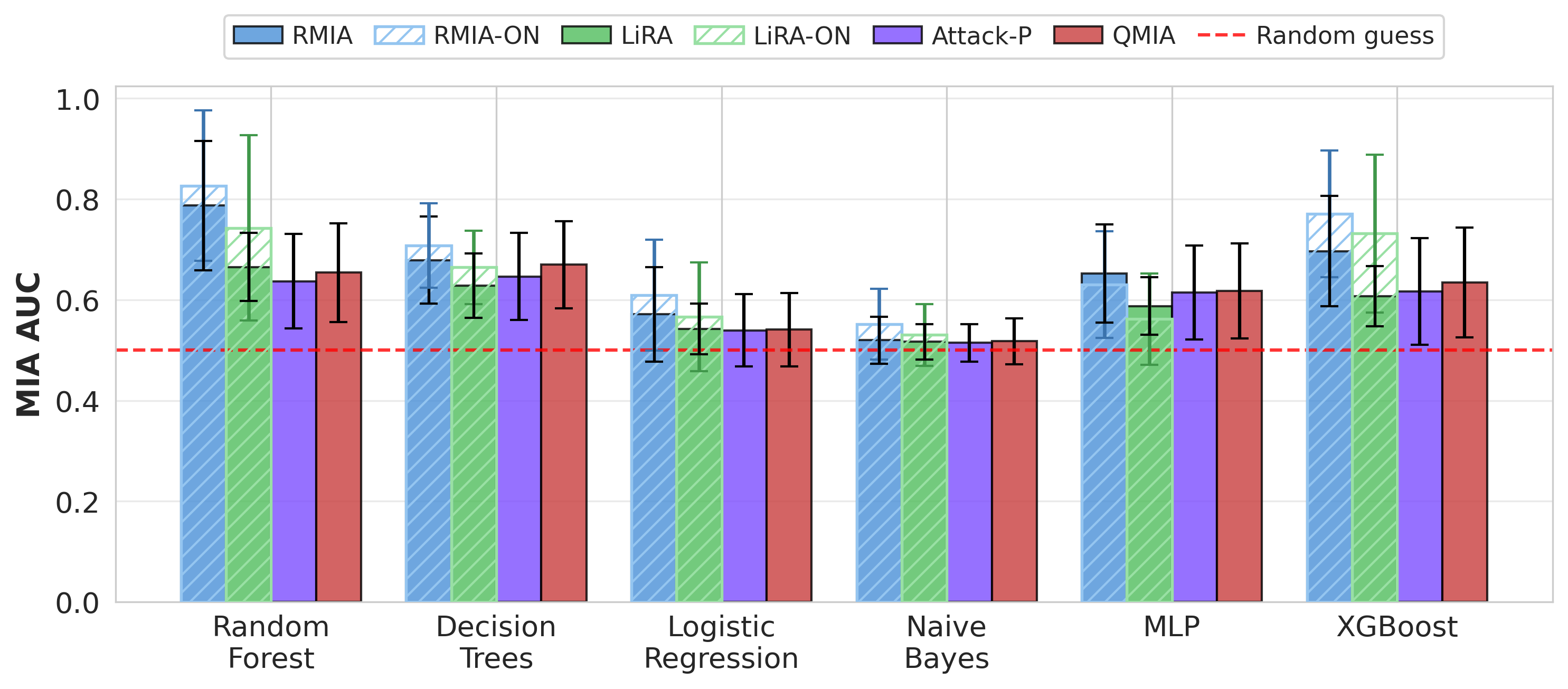}
    \caption{MIA AUC results for all attacks and models, averaged across all datasets.}
    \label{fig:all-res-barplot}
\end{figure}

General attacks' performance is present in Figure~\ref{fig:all-res-barplot}. We observe that RMIA dominates all the remaining attacks. We also teste online version of LiRA and RMIA showing small improvements, expect for LiRA in RF and XGB models.

\section{Reference models on RMIA}~\label{app:ref_models}

\begin{figure}[ht!]
    \centering
    \includegraphics[width=\columnwidth]{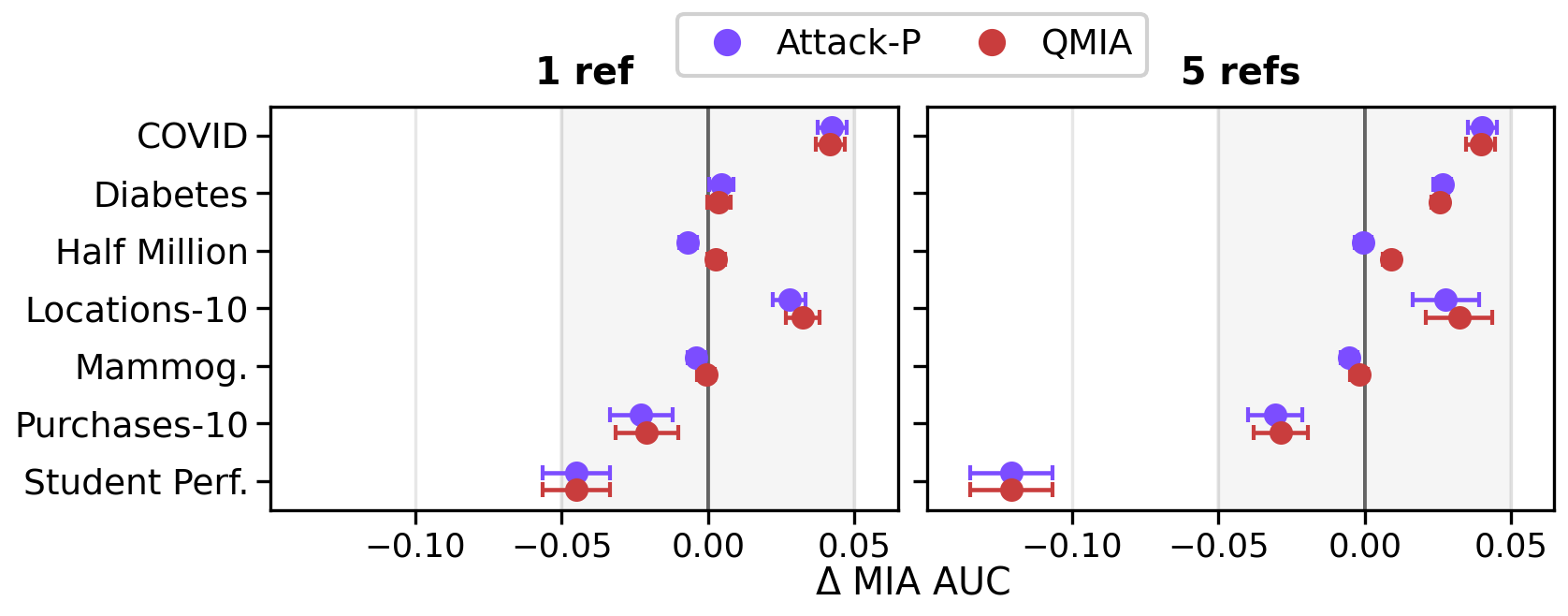}
    \caption{Reference model equivalence gap for surrogate RMIA (1 and 5 reference models) versus Attack-P/QMIA. Each dataset point shows the mean MIA AUC gap, with the error gaps representing the Standard Error Mean across surrogate models. The shaded band denotes the 0.05 margin used in the TOST equivalence test. }
    \label{fig:proxy-extension-all-ds}
\end{figure}

Figure~\ref{fig:proxy-extension-all-ds} extends the comparison of attack performance between surrogate RMIA and Attack-P/QMIA for all datasets, via TOST equivalence test. Here, we provide the results for 1 and 5 reference models since there was little improvement in comparing with more models, as stated in Section~\ref{sec:reference_models}. Six datasets stay within the equivalence margin band, highlighting the equivalence of RMIA versus reference-model free attacks. In contrast, this does not fully happen for \textit{Student Performance}, especially in the 5 reference-model case, likely because it is a small and overfit-prone dataset.

\section{Comparison of single-out MIAs}~\label{app:singleouts}



\begin{figure}[ht!]
    \centering
    \includegraphics[width=\columnwidth]{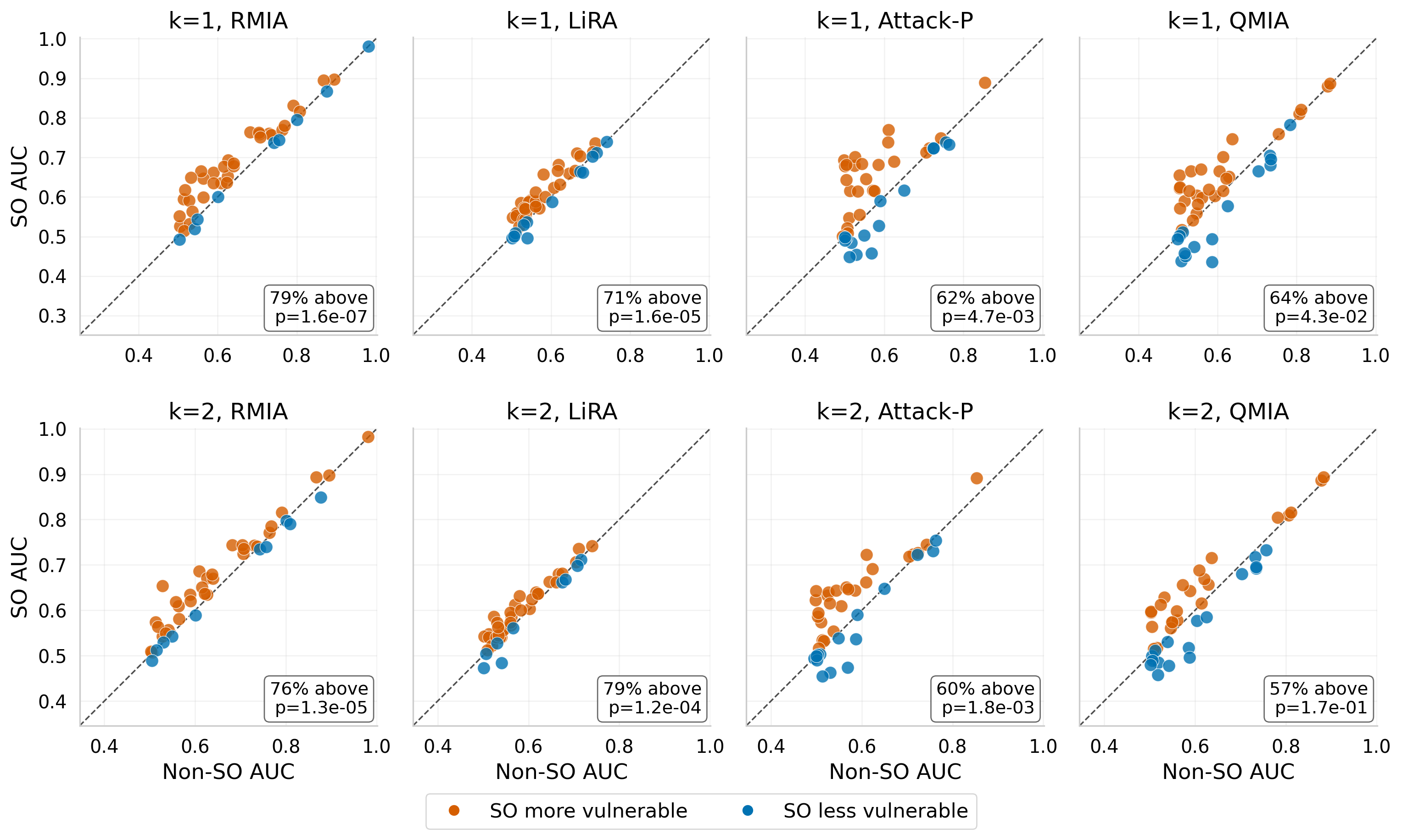}
    \caption{SO vs. non-SO attack AUC, faceted by attack and anonymity threshold $k$.}
    \label{fig:singleoutsMIA_attacks}
\end{figure}

To assess how each model identifies SO records, Figure~\ref{fig:singleoutsMIA_attacks} compares mean AUCs for SO and non-SO records, with each point representing a dataset-model pair averaged. The box shows the percentage of pairs where SO AUC exceeds non-SO AUC and the one-sided paired Wilcoxon signed-rank $p$-value testing whether SO AUC is systematically higher.

Across panels, most points lie above the diagonal, indicating that SO records are generally more vulnerable to MIAs than non-SO records. 
The strength of this effect also depends on the attack. 
RMIA and LiRA form tight bands along the diagonal, with SO AUC consistently slightly higher than non-SO AUC across the full range of attack performance, indicating a uniform and reproducible SO vulnerability effect across dataset/model combinations. In contrast, Attack-P and QMIA show substantially greater dispersion, particularly at lower AUCs (0.5–0.65). 
QMIA shows the weakest and least reliable SO effect, which further weakens at $k=2$, where it no longer reaches significance and its win rate approaches chance.


\section{Effectiveness of defenses over all datasets}~\label{app:defenses}
\begin{figure}[hbt]
    \centering
    \includegraphics[width=\columnwidth]{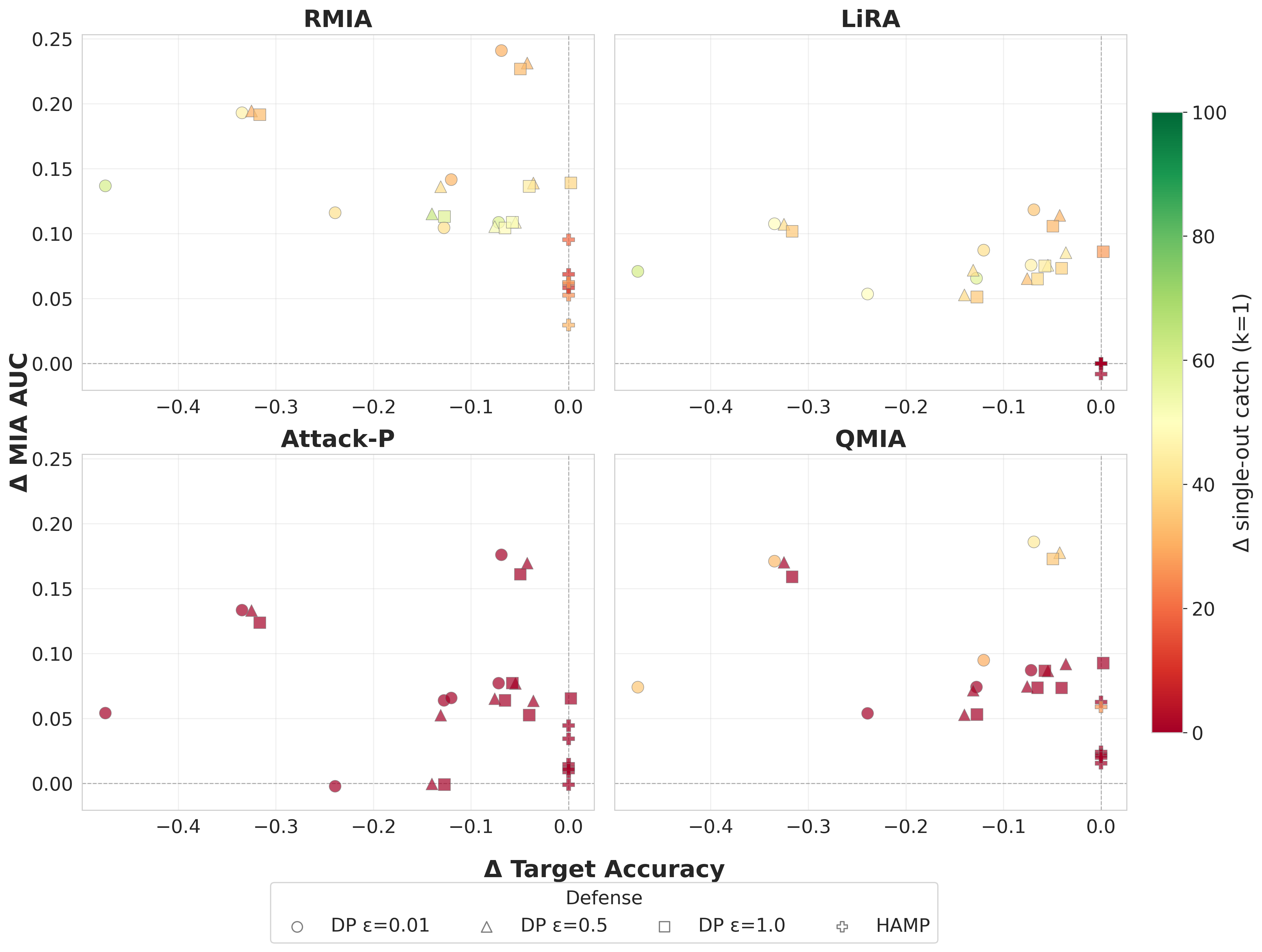}
    \caption{Resulting application of defenses on the catch of single-outs for k=1. Each point represents a dataset-defense combination, averaging across the 6 models.}
    \label{fig:singleouts-defenses-k1}
\end{figure}

Figure~\ref{fig:singleouts-defenses-k1} shows the impact of defenses on all attacks. RMIA and LiRA present a greater drop in single-out catch, following the MIA AUC reduction. For the reference-model free approaches, the drop in attack success does not translate to reduced vulnerability of single-outs.

\end{document}